%% file: a.tex
\documentclass[a4paper]{article}
\usepackage{amsmath, amssymb}
\usepackage{epsfig}
\usepackage{comment}
\usepackage[backend=bibtex, style=ieee, citestyle=numeric-comp]{biblatex}
\usepackage{floatflt}
\usepackage{epstopdf}
\usepackage{glossaries}
\usepackage{listings}
\usepackage{chngcntr}
\usepackage{diagbox}
\usepackage{nicematrix}
\usepackage{multirow}
\usepackage{wrapfig}
\usepackage[font=small,labelfont=bf]{caption}
\usepackage{tabularx}
\usepackage{tikz}
\usetikzlibrary{positioning, calc, shapes.geometric, fit, backgrounds, 3d}
\usepackage[colorlinks=true, linkcolor=black, urlcolor=magenta, citecolor=black]{hyperref}
\usepackage{enumitem, kantlipsum}
\usetikzlibrary{shapes, arrows.meta, positioning}
\usepackage{mdframed}
\usepackage{ifthen}
\makeglossaries
\usepackage{fancyhdr}
\usepackage{parskip}
\usepackage{csquotes}
\usepackage{caption}
\usepackage{subcaption}
\usepackage{algorithm}
\usepackage{algpseudocode}
\usepackage[top=3cm, bottom=3cm,inner=3cm, outer=3cm]{geometry}

\DeclareMathOperator{\EX}{\mathbb{E}}

\usepackage{eso-pic}								

\newcommand{\ARBETE}{ 
 \newline \noindent Kandidatarbete inom civilingenjörsutbildningen vid Chalmers 
 \medskip
}

\newcommand{\titel}{ClaudesLens: Uncertainty Quantification in Computer Vision Models} 
\newcommand{\undertitel}{} 
\newcommand{\engtitel}{ClaudesLens: Osäkerhetskvantifiering hos datorseende modeller} 

\newcommand{\namn}{ 
  Mohamad Al Shaar \\
  Nils Ekström  \\
  Gustav Gille \\
  Reza Rezvan \\
  Ivan Wely
}

\newcommand{\examina}{ 
 \newline \noindent {\it Kandidatarbete i matematik inom civilingenjörsprogrammet Automation och Mekatronik vid Chalmers} \smallskip
 \newline \noindent Mohamad Al Shaar
 \bigskip
 \newline \noindent {\it Kandidatarbete i matematik inom civilingenjörsprogrammet Datateknik vid Chalmers} 
 \smallskip
 \newline \noindent Nils Ekström
 \quad Reza Rezvan
 \quad Ivan Wely
 \bigskip
   \newline \noindent {\it Kandidatarbete i matematik inom civilingenjörsprogrammet Informationsteknik vid Chalmers} \smallskip
   \newline \noindent Gustav Gille
   \bigskip
 \bigskip

}

\newcommand{\handledare}{
Moritz Schauer
}

\newcommand{\skribenter} {\begin{tabular}{l} \namn \end{tabular}}

\newcommand{\omslag}{
\newgeometry{top=3cm, bottom=4cm,left=2 cm,right=1cm}	
\pagestyle{fancy}
\pagenumbering{gobble}
\fancyhead[C]{\includegraphics[width=170mm]{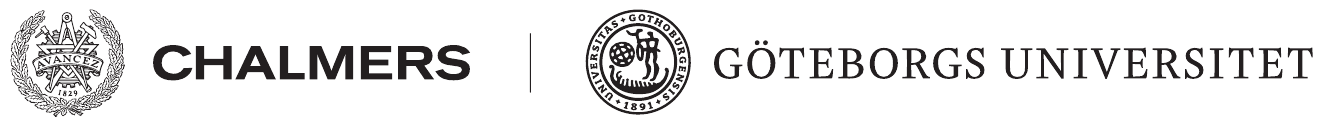}\\}
\addtolength{\voffset}{0.3cm}
\renewcommand{\headrulewidth}{1pt}

\parbox{17cm}{
\vspace{60mm}

\noindent{\Huge \titel}
\bigskip

\noindent {\Large \undertitel}
\bigskip

\noindent{\huge \engtitel}

\noindent\hspace*{-1 ex}{\Large \it
\ARBETE
}
\vspace{20mm}

\noindent\hspace*{-1 ex}\parbox{80mm}{\noindent {\huge \skribenter}}}

\renewcommand{\footrulewidth}{1pt}

\fancyfoot[L]{\vspace{0.1mm}\large Institutionen för Matematiska vetenskaper\\
CHALMERS TEKNISKA HÖGSKOLA\\
GÖTEBORGS UNIVERSITET\\
Göteborg, Sverige 2024}

\newpage
\thispagestyle{empty}
\mbox{}

\newpage}

\newcommand{\titelsidor}{
\newgeometry{top=3cm, bottom=5cm,left=3 cm,right=3cm}

\thispagestyle{fancy}
\fancyhf{}
\renewcommand{\headrulewidth}{0pt}

\mbox{}
\vspace{50mm}

\noindent {\LARGE \titel}\bigskip\bigskip

\noindent {\large \undertitel}
\vspace{30mm}

\noindent {\large \examina}

\vfill
\hspace{-2.4 ex} \begin{tabular}[t]{lll}
Handledare:& \handledare & \end{tabular}

\renewcommand{\footrulewidth}{0pt}

\fancyfoot[L]{\vspace{0.1mm}\large Institutionen för Matematiska vetenskaper\\
CHALMERS TEKNISKA HÖGSKOLA\\
GÖTEBORGS UNIVERSITET\\
Göteborg, Sverige 2024 }

\newpage
\thispagestyle{fancy}
\fancyhf{}
\newgeometry{top=3cm, bottom=3cm,left=3 cm,right=3cm}
\mbox{}
\vfill

 \setcounter{page}{0}
\newpage }

\usepackage[T1]{fontenc}                
\usepackage[swedish,english]{babel}             
\pdfoutput=1
\begin{document}
\omslag
\titelsidor
\thispagestyle{empty}
\newgeometry{top=3cm, bottom=3cm,left=3 cm,right=3cm}

\section*{Preface} \label{sec:preface}
\input{sections/Preface/preface}
\newpage

\section*{Popular Science Presentation} \label{sec:presentation}
\input{sections/Presentation/presentation}
\newpage

\selectlanguage{swedish}
\input{sections/sammandrag}

\selectlanguage{english}
\input{sections/abstract} 
\newpage

\pagestyle{plain}
\newpage

\tableofcontents
\newpage

\pagenumbering{arabic}
\section{Introduction} \label{sec:intro}
\input{sections/Introduction/introduction}

\counterwithin{equation}{section}
\renewcommand{\theequation}{2.\arabic{equation}}
\section{Background} \label{sec:background}
There are many different and diverse concepts related to machine learning, computer vision, and uncertainty. The following section will describe the most essential parts used in our research. The explanations are not exhaustive, but should rather give the reader an understanding of how they work and their purposes.

\input{sections/Background/NN}
\input{sections/Background/CV}
\input{sections/Background/CNN}
\input{sections/Background/ConvNext}
\input{sections/Background/ViT}
\input{sections/Background/UQ}

\counterwithin{equation}{section}
\renewcommand{\theequation}{3.\arabic{equation}}
\section{Method} \label{sec:method}
This section lays the foundation for the theoretical framework, how the proposed models were implemented, what tools were used, and how the evaluation was conducted.

\input{sections/Method/hypothesis}
\input{sections/Method/dataset}
\input{sections/Method/pertubation_injection}
\input{sections/Method/training_implementation}

\input{sections/Method/Models/naive}
\input{sections/Method/Models/ConvNext}
\input{sections/Method/Models/ViT}
\renewcommand{\theequation}{4.\arabic{equation}}
\section{Results} \label{sec:result}
\input{sections/Results/results}

\section{Discussion} \label{sec:discussion}
\input{sections/Discussion/discussion}

\section{Societal and Ethical Aspects} \label{sec:soc_eth}
\input{sections/Ethics/Ethics}

\newpage

\printbibliography
\clearpage

\appendix
\pagenumbering{roman}
\counterwithin{equation}{section}
\section{Appendix -- Source Code}\label{appendix:c}
\input{appendix/C}
\end{document}

%% file: sections/Preface/preface.tex
This bachelor’s thesis has been carried out at the Department of Mathematical Sciences at Chalmers
University of Technology. We would like to thank our supervisor Moritz Schauer for his dedicated commitment and guidance during the work.

The group believes that all members have contributed equally to the project. All group members’ work has been recorded in a logbook. This logbook is not included here but below the group members’ contributions to each section of the text in the report are presented.

\begin{table}[H]
    \centering
    \caption{Contribution Report}
    \begin{tabular*}{\textwidth}{l@{\extracolsep{\fill}}lll}
    \hline
    \hline
    \textbf{Chapter} & \textbf{Title} & \textbf{Author(s)} \\
    \hline
            & Preface & Reza \\
            & Popular science presentation & Nils, Reza \\
            & Sammandrag & Nils, Reza \\
            & Abstract & Nils, Reza \\
    \hline
    1       & Introduction & Mohamad, Reza, Ivan \\
    1.1     & Purpose & Reza \\
    1.2     & Objectives & Gustav \\
    1.3     & Scope & Reza \\
    \hline
    2     & Background & Mohamad, Reza, Ivan \\
    2.1   & Neural Networks & Mohamad, Reza \\
    2.1.1 & Structure of a Neuron & Gustav, Reza \\
    2.1.2 & Notation for Entire Neural Networks & Gustav, Reza, Ivan \\
    2.1.3 & Training and Loss Function & Mohamad, Reza \\
    2.1.4 & Backpropagation & Ivan, Gustav \\
    2.2 & Computer Vision & Gustav \\
    2.3 & Convolutional Neural Networks & Gustav, Ivan \\
    2.3.1 & Convolutional Layers & Gustav, Ivan \\
    2.3.2 & Layer Normalization & Mohamad, Gustav, Ivan \\
    2.3.3 & Pooling & Gustav \\
    2.3.4 & ConvNeXt & Gustav, Ivan \\
    2.4 & Vision Transformers & Gustav \\
    2.4.1 & Attention & Mohamad, Nils, Gustav \\
    2.4.2 & Embedding & Gustav \\
    2.4.3 & Multiheaded Self-Attention & Gustav \\
    2.5 & Uncertainty in Information Theory & Reza, Ivan, Gustav \\
    \hline
    3 & Method & Mohamad, Reza, Ivan \\
    3.1 & Entropy-based Uncertainty Quantification Framework & Nils, Reza, Ivan \\
    3.1.1 & Perturbation in Neural Networks & Nils, Reza, Ivan \\
    3.1.2 & Distribution of Perturbed Matrices & Mohamad, Nils, Reza, Ivan \\
    3.1.3 & PI: Perturbation Index & Nils, Reza, Ivan \\
    3.1.4 & PSI: Perturbation Stability Index & Everyone \\
    3.2 & Technical Choices and Dataset &  Mohamad, Reza\\
    3.3 & Perturbation Injection & Gustav, Reza\\
    3.4 & Training and Implementations of Models & Mohamad, Reza\\
    3.4.1 & Na\"ive Multinomial Regression & Gustav\\ 
    3.4.2 & ConvNeXt Model &  Gustav\\
    3.4.3 & Vision Transformer Model & Mohamad, Gustav\\
    \hline
    4 & Results & Everyone \\
    4.1 & Na\"ive Multinomial Regression & Nils, Reza, Ivan \\
    4.2 & Pretrained ConvNeXt & Nils, Reza, Ivan, Gustav \\
    4.3 & Pretrained ViT & Mohamad, Nils, Reza, Ivan \\
    \hline
    5 & Discussion & Everyone \\
    5.1 & Future Improvements & Everyone \\
    5.2 & Conclusion & Everyone \\
    \hline
    6 & Societal and Ethical Aspects & Mohamad \\
    6.1 & Accountability and Impact on Decision Making & Mohamad \\
    \hline
    A & Appendix -- Source Code & Ivan\\
    B & Appendix -- Graphs & Nils\\
    \hline
    \hline
    \end{tabular*}
    \label{tab:contribution}
\end{table}

%% file: sections/Presentation/presentation.tex
Vision and the ability to identify objects are fundamental in our lives and are incredibly complex concepts and processes. When we are born, we do not understand the concepts of shapes, animals, or their inherent meanings. Young children have a difficult time distinguishing a cat from a dog. They gradually learn what distinguishes a dog from a cat by pointing out and labeling animals, and through trial and error, they grasp what features define a dog. This is intelligence; perceive, analyze, and draw conclusions. Artificial intelligence refers to the ability of \textit{machines} to perceive, analyze, and draw conclusions on their own. 

Vision is a complex human sense and an obvious step in developing machine intelligence would therefore be to replicate a vision system. Neural networks are like children who are new to the world and have to learn what features distinguish dogs from cats.

Neural networks learn much like children do. The neural net takes in data, analyzes it thoroughly, and arrives at a final conclusion. In the same way that you point out how a cat is not a dog to a child, you tell the neural network how \textit{far} it is from the correct answer, in hopes that it will do better on its next attempt. The neural network will slightly change its approach to the problem and try again. This process is repeated until the predictions start to make sense. However, both children and neural networks can be adamant about their way of learning. How do we \textit{really} know if they've understood what a dog is? Can we find a method to quantify and evaluate the \textit{uncertainty} they may have?

\begin{figure}[H]
    \centering
    \includegraphics[width=0.7\textwidth]{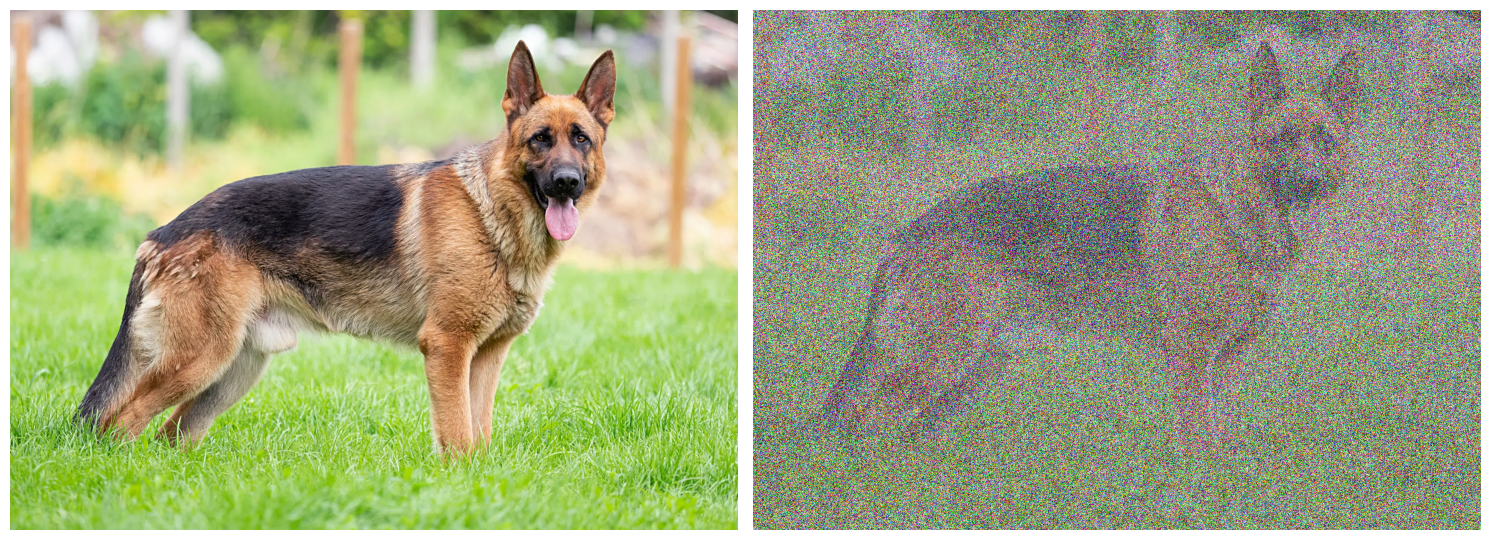}
    \caption{A picture of a dog and the same image but perturbed.}
    \label{fig:dawg}
\end{figure}

We believe that we can determine that they have properly understood the concept of a dog by showing a noisy picture of a dog. If they still can determine that the picture is that of a dog, they might have understood the characteristics of a dog. Not only this but if they still can determine the picture of the dog, can we \textit{predict} what they would say if we presented them with a \textit{new} picture? This aspect of testing can be generalized beyond just the image, as we can put the neural network in a noisy environment and see if it can still make correct predictions.

We hypothesize that by putting a neural network in a noisy environment, which we call "shaking" the network, one can understand if the network has properly understood the concepts of the problem and how it would perform on new similar data. What happens with the output of the network? Especially, what happens with networks which are designed to see, so-called \textit{computer vision models}. 

The results from this study show that this is indeed a way to quantify and evaluate the uncertainty of the outputs from different networks. Our theoretical framework and methods have been tested on three different models where we draw logical conclusions based on our theory. Our methods might not be the most rigid and absolute,
but we believe that the foundation can be built upon in today's world of artificial intelligence. AI and its influence are certainly not going to stop anytime soon. Finding ways to quantify the uncertainty of models' predictions and how their performance would be on new data, before even seeing it, is a powerful tool in an uncertain world.

%% file: sections/sammandrag.tex
\renewcommand{\abstractname}{Sammandrag}
\begin{abstract} \label{sec:sammandrag}
I en värld där allt fler beslut bestäms med hjälp av artificiell intelligens är det av yttersta vikt att säkerställa att dessa beslut är välgrundade. Neurala nätverk är de moderna byggstenarna för artificiell intelligens. Moderna modeller för datorseende baserade på neurala nätverk används ofta för objektklassificering. Att korrekt klassificera objekt med \textit{säkerhet} har blivit allt viktigare på senare tid. Att kvantifiera \textit{osäkerhet} i neuralt nätverks utdata är dock en utmanande uppgift. Här visar vi en möjlig metod för att kvantifiera och utvärdera osäkerheten i utdatan från olika datorseendemodeller baserat på Shannon-entropi. Genom att lägga till störningar på olika nivåer, på olika delar; allt från indatan till parametrarna i nätverket, kan man introducera entropi i systemet. Genom att kvantifiera och utvärdera de störda modellerna med de föreslagna PI- och PSI-måtten kan vi dra slutsatsen att vårt teoretiska ramverk kan ge insikt i osäkerheten i utdatan för datorseendemodeller. Vi tror att detta teoretiska ramverk kan användas för olika tillämpningar inom neurala nätverk. Vi tror att Shannon-entropi kan så småningom ha en större roll i SOTA-metoderna (State-of-the-art) för att kvantifiera osäkerhet inom artificiell intelligens. En dag kanske vi kan tillämpa Shannon-entropi på våra neurala system.
\end{abstract}

%% file: sections/abstract.tex
\begin{abstract} \label{sec:abstract}
In a world where more decisions are made using artificial intelligence, it is of utmost importance to ensure these decisions are well-grounded. Neural networks are the modern building blocks for artificial intelligence. Modern neural network-based computer vision models are often used for object classification tasks. Correctly classifying objects with \textit{certainty} has become of great importance in recent times. However, quantifying the inherent \textit{uncertainty} of the output from neural networks is a challenging task. Here we show a possible method to quantify and evaluate the uncertainty of the output of different computer vision models based on Shannon entropy. By adding perturbation of different levels, on different parts, ranging from the input to the parameters of the network, one introduces entropy to the system. By quantifying and evaluating the perturbed models on the proposed PI and PSI metrics, we can conclude that our theoretical framework can grant insight into the uncertainty of predictions of computer vision models. We believe that this theoretical framework can be applied to different applications for neural networks. We believe that Shannon entropy may eventually have a bigger role in the SOTA (State-of-the-art) methods to quantify uncertainty in artificial intelligence. One day we might be able to apply Shannon entropy to our neural systems.
\end{abstract}

%% file: sections/Introduction/introduction.tex
Neural networks, which mimic the human brain's structure \cite{Rosenblatt_1957_6098}, have advanced technology by learning from data and adapting autonomously \cite{rumelhart1986learning}. They are now central to artificial intelligence, with significant applications in computer vision, where they enable computers to recognize objects in images and videos \cite{Fukushima1980}.

However, the increasing use of neural networks brings challenges, especially in uncertainty quantification. Traditionally, this involves analyzing and predicting errors to ensure reliable outputs, uncertainty can be defined in terms of Shannon entropy \cite{shannon1948mathematical}. In artificial intelligence, uncertainty quantification is crucial due to the probabilistic nature of learning and factors like noisy data, unmet model assumptions, and inherent randomness in sampling methods. These issues can compromise the real-world performance of models that appear accurate in controlled tests.

\subsection{Purpose} \label{sec:purpose}
The purpose of this paper is to explore methods that evaluate uncertainty, using Shannon entropy, in machine learning models, with a focus on computer vision models.

\subsection{Objectives} \label{sec:objectives}
This paper will investigate the following questions:
\begin{enumerate}
    \item Can entropy be used to find methods for quantifying the uncertainty of a machine learning model?
    
    \item If this is possible, in what ways can these methods be applied?
    
    \item Does this work equally well on all models?
\end{enumerate}

\subsection{Scope} \label{sec:scope}
To examine and evaluate these methods, three different models will be tested. The models are the following:

\begin{enumerate}
    \item A na\"ive multinomial regression model
    \item A Convolutional neural network (CNN) model, base ConvNeXt.
    \item The original vision transformer model, base ViT. 
\end{enumerate}

%% file: sections/Background/NN.tex
\subsection{Neural Networks}\label{Neural Network} \label{sec:neural_network}
Neural networks are the building blocks of modern machine learning, inspired by biological neural networks \cite{Rosenblatt_1957_6098}. To effectively solve problems, a neural network undergoes two phases: a training phase, where the network \textit{learns} from a dataset, and the prediction phase, where the network applies its learned patterns to unseen data. This section will define the structure and operational processes of neural networks.

\subsubsection{Structure of a Neuron} \label{sec:nn_structure_neuron}
At the heart of the neural network lies the \textit{neuron}, modeled after its biological counterpart \cite{Rosenblatt_1957_6098}. Each neuron consists of one or more inputs $x_j$ and a single output $y$. All inputs are scaled by an associated weight $w_j$,  and the neuron's output, $y$, is determined by these inputs and weights. The core functionality of a neuron is divided into two main components: the summation function and the activation function (see Figure~\ref{fig:neuron} for an overview). 

\input{imgs/latexpaint/neuron}

Each neuron has a special input, the \textit{bias} $b$ which adjusts the neuron's activation threshold. It is also commonly denoted as $w_0$. Inputs feed into the summation function,

\begin{equation}\label{eq: weighted sum}
    z = \sum_{j=1}^{n} w_{j} x_{j} + b, \; \; b=w_{0},
\end{equation}

which computes the weighted summation of the inputs and the bias.

\begin{figure}[H]
    \centering    \includegraphics[width=0.6\textwidth]{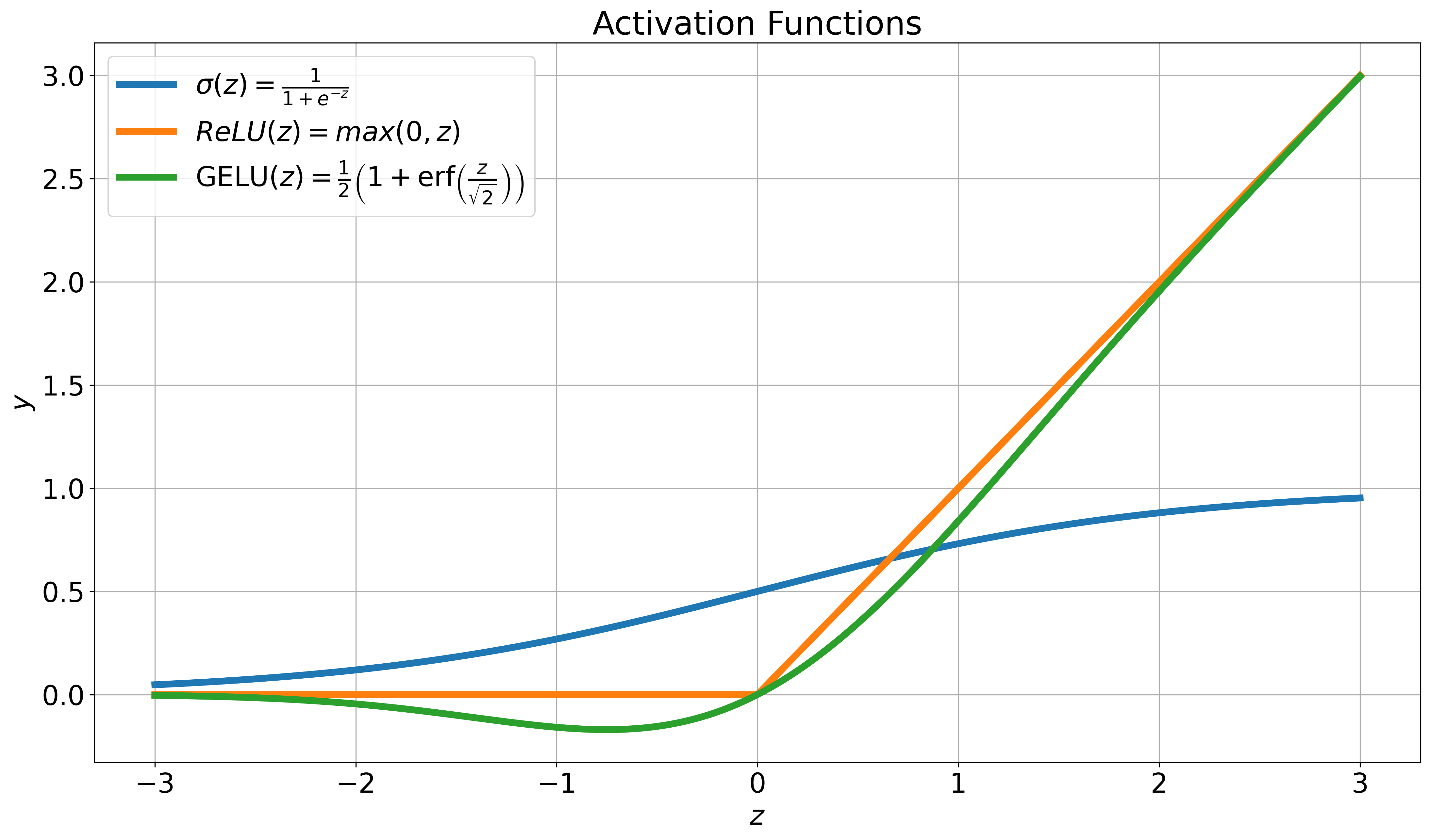}
    \caption{The sigmoid, ReLU, and GELU activation functions and their graphical representations.}
    \label{fig:activation_functions}
\end{figure}

The activation function, denoted as $f$, calculates the neuron's output $y = f(z)$ based on the weighted summation. Activation functions such as the sigmoid, ReLU, or GELU (see Figure~\ref{fig:activation_functions}) introduce non-linearity, enabling neural networks to approximate complex, non-linear functions \cite{aktiveringsfunktioner}. This capability is crucial for capturing the intricate patterns present in real-world data.

\subsubsection{Notation for Entire Neural Networks} \label{sec:nn_notation}
Vector and matrix notations offer a concise and efficient representation of neural network operations which are beneficial for networks with numerous inter-neuron connections.

Consider a neuron's inputs represented as a vector $\mathbf{x} = \left[x_1, x_2, \ldots, x_n\right]$, where each $x$ corresponds to an input. The associated weights for the neuron are similarly denoted by vector $\mathbf{w} = \left[w_{1}, w_{2}, \ldots, w_{n}\right]$.

The summation function's operation simplifies to the dot product of the two vectors and the addition of the bias, computed using the dot product,

\begin{equation}
    z = \mathbf{w} \cdot \mathbf{x} + b.
\end{equation}

Now consider multiple neurons with numerous inter-neuron connections with multiple outputs (see Figure~\ref{fig:multiple_neurons} for an example). The \textit{layers} between the input neurons and the output neurons are called \textit{hidden layers}. A \textit{fully connected layer} is a layer where each neuron connects to every neuron in the preceding layer. The matrix-vector equation, 

\begin{equation}
    \mathbf a = \mathbf{W}_1 \mathbf{x} + \mathbf{b}_1 = \left[a_1, a_2, \ldots, a_n\right],
\end{equation}

yields each output of each neuron in the hidden layer. 

$\mathbf{W}_1$ is the \textit{weight matrix} with rows $\mathbf{w}_i = \left[w_{i, 1}, w_{i, 2}, \ldots w_{i, j}, \ldots, w_{i, n} \right]$ where each element corresponds to the $i$-th neuron and the $j$-th weight. The bias values in the hidden layer are represented by the vector $\mathbf{b}_1 = \left[b_1, b_2, \ldots, b_n\right]$. In the case of several layers of neurons, we work with several weight matrices and bias vectors, indexed with $\mathbf{W}_l$, and $\mathbf{b}_l$ respectively, for each layer $l \in [1, \dots, L]$. 

\begin{equation}
    \text{Input}: \mathbf{x} \quad
    \text{Hidden Layer}: \mathbf{a} = f.(\mathbf{W}_1 \mathbf{x} + \mathbf{b}_1) \quad
    \text{Output Layer}: \mathbf{y} = f.(\mathbf{W}_2 \mathbf{a} + \mathbf{b}_2)
\end{equation}

Where the dot indicates that the function acts on each element of the vector, this is henceforth the functionality of all activation functions.

\input{imgs/latexpaint/neural_network_1_output}

Important to note is that activation functions across layers can differ. A common practice is to use a certain activation function (for example ReLU) for all hidden layers and use different activation function at the output layer. In problems where the final answer should be interpreted as probabilities, the softmax function is commonly used at the output layer. In this paper, the softmax function is used to transform the output into a probability vector that sums up to $1$ and keeps the vector ordering. Henceforth we will assume that the output of a neural network will be a probability vector.

\subsubsection{Training and Loss Function} \label{sec:nn_train_loss}
In machine learning, models are trained to learn from data and generalize, aiming to minimize discrepancies between predictions and actual outcomes during the training phase. This discrepancy is measured using a \textit{loss function}, with the primary goal being to minimize this loss to enhance accuracy, which is defined as the fraction of correct predictions. For classification tasks, the commonly used \textit{Average Cross-Entropy Loss} is defined as, 

\begin{equation}
    L(y, \hat{y}) = - \frac{1}{N} \sum_{i=1}^N \sum_{j=1}^C y_{i,j} \log(\hat{y}_{i,j}) = -\sum_{i = 1}^N\EX_{j \sim y_i} \left[ \log(\hat{y}) \right],
\end{equation} 
 
where $y_{i,j}$ is 1 if class $j$ is the correct classification for sample $i$ and 0 otherwise, and $\hat{y}_{i,j}$ is the predicted probability that sample $i$ belongs to class $j$. This loss function computes the logarithmic loss averaged over all samples and classes. This can be simplified into an expected value. We can average the logarithmic loss over all classes weighted by the true distribution $y$.

\subsubsection{Backpropagation} \label{sec:nn_backprop}
The backpropagation algorithm employs the loss function to determine how the network should minimize the error. It does so by using the chain rule with the loss function's gradient with respect to each weight/bias in each neuron in the neural network \cite{rumelhart1986learning}.
This partial derivative provides the direction and magnitude of the \textit{parameter's} deviation from the loss function, which can be used to minimize it. A parameter of a neural network is a variable that can be optimized in this fashion. The algorithm is as follows:

\begin{enumerate}    
    \item \textbf{Gradient Calculation}: Computing the gradient of the loss function $L$ for each neuron's weight $\mathbf{w}$ and bias $b$ by applying the chain rule in reverse, from the output layer back to the input layer: $\partial L / \partial w_{ij}^{(l)}, \; \; \partial L / \partial b_{i}^{(l)}$ For the layer index $l$, neuron index $i$, and weight index $j$. 
    
    \item \textbf{Weight Update}: Adjusting the weights using an optimization algorithm (for example the Adam optimizer \cite{AdamOptimizer}), typically in the opposite direction of the loss function's gradient with respect to the variable in context to minimize the error: 
    $\Delta w = - \eta \frac{ \partial L}{\partial w^{(l)}_{ij}}$, where $eta$ is the scaling commonly known as the \textit{learning rate.}
    
    \item \textbf{Repetition}: This optimization algorithm is applied over the entire dataset, henceforth denoted as $\mathcal{D}$. One iteration of this is called an \textit{epoch}. This process is \textit{repeated} for some amount of epochs or if a certain metric, such as accuracy, or loss has met a criterion.
\end{enumerate}

In order to know how the loss function $L$ is impacted by each parameter in the network, for example, $w^{(l)}_{ij}$, it must be known how the loss function changes with respect to said parameter. This can be calculated as,

\begin{equation}
    \begin{aligned}
    \frac{\partial{L}}{\partial w_{ij}^{(l)}} = 
    \frac{\partial{L}}{\partial{ \hat{y}_i^{(l+1)}}}
    \frac{\partial \hat{y}_i^{(l+1)}}{\partial w^{(l)}_{ij}} =     
    \frac{\partial{L}}{\partial{ \hat{y}_i^{(l+1)}}}
    \frac{\partial \hat{y}_i^{(l+1)}}{\partial z_i^{(l+1)}}
    \frac{\partial z_i^{(l+1)}}{\partial w^{(l)}_{ij}}
    \end{aligned},
\end{equation}

where $z$ is the input to the activation function $y$ at layer $l+1$, which consists of the weighted summation from the previous layer $l$. 

By calculating the expanded equation, the partial derivative for parameter $w_{ij}^{(l)}$ is obtained which is utilized as, 
\begin{equation}
    \Delta w = -\eta \frac{\partial L}{\partial w^{(l)}_{ij}}.
\end{equation}

By finding the gradient, the variable can be nudged in the opposite direction (gradient descent) to minimize the loss function. 

%% file: imgs/latexpaint/neuron.tex
\begin{figure}[H]
\begin{center}
\scalebox{0.7}{
\begin{tikzpicture}[
    node distance=1cm,
    line width=1pt,
    >={Stealth[length=3mm]},
    neuron/.style={circle, draw, minimum size=1cm},
    act_func/.style={rectangle, draw, minimum height=1cm, minimum width=1.5cm},
    layer/.style={circle, draw, minimum height=1cm, minimum width=1.5cm,},
    annot/.style={text width=4cm, align=center, anchor=north}
]

\node[neuron]                   (input1) {$x_1$};
\node[neuron, below=of input1]  (input2) {$\ldots$};
\node[neuron, below=of input2]  (input3) {$x_n$};

\node[layer, right=2cm of input2] (sum) {$\Sigma$};
\node[neuron, above=of sum] (bias) {$b$};

\node[act_func, right=2cm of sum] (f) {$f$};

\node[neuron, right=2cm of f] (y) {$y$};


 
\draw[->] (input1) -- (sum) node [midway, above, sloped] (TextNode) {$w_1$};
\draw[->] (input2) -- (sum) node [midway, above, sloped] (TextNode) {$w_j$ };
\draw[->] (input3) -- (sum) node [midway, above, sloped] (TextNode) {$w_n$};
 \draw[->] (bias) -- (sum);
\draw[->] (sum) -- (f);
\draw[->] (f) -- (y);
\draw[->] (sum) -- (f) node [midway, above, sloped] (TextNode) {$(z)$};

\node[annot, above=of input1]  (input)  {Inputs};
\node[annot, right=-1 of input] (summation)  {Summation};
\node[annot,  right=-0.8 of summation]  (act_func){Activation function};
\node[annot,  right=-0.90 of act_func]  {Output};

\end{tikzpicture}}
\caption{A single neuron with $n$ inputs and one output, showcasing the summation and activation components in a neuron.} 
\label{fig:neuron}
\end{center}
\end{figure}

%% file: imgs/latexpaint/neural_network_1_output.tex
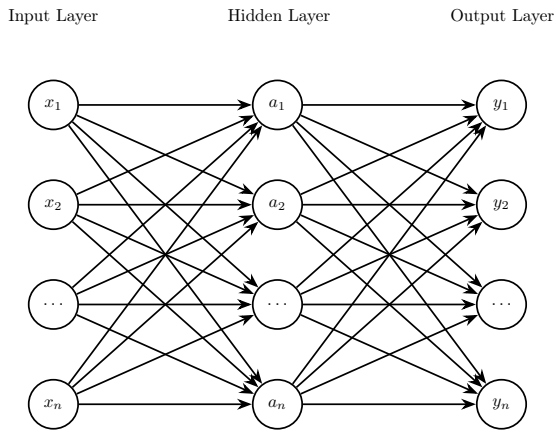
\begin{figure}[H]
\centering
\scalebox{0.65}{
    \begin{tikzpicture}[
        node distance=1cm,
        line width=1pt,
        >={Stealth[length=3mm]},
        neuron/.style={circle, draw, minimum size=1cm},
        layer/.style={circle, draw, minimum size=1cm},
        annot/.style={text width=4cm, align=center, anchor=north}
    ]
    
    \node[annot, above=of input1]  (input layer)  {Input Layer};
    \node[annot, right=0.3cm of input layer] (hidden layer)  {Hidden Layer};
    \node[annot,  right=0.25cm of hidden layer] (output layer)  {Output Layer};
    
    \node[neuron]                   (input1) {$x_1$};
    \node[neuron, below=of input1]  (input2) {$x_2$};
    \node[neuron, below=of input2]  (input3) {$\ldots$};
    \node[neuron, below=of input3]  (input4) {$x_n$};
    
    \node[layer, right=3.5cm of input1] (layer1) {$a_1$};
    \node[layer, below=of layer1] (layer2) {$a_2$};
    \node[layer, below=of layer2] (layer3) {$\ldots$};
    \node[layer, below=of layer3] (layer4) {$a_n$};
    
    \node[neuron, right=3.5cm of layer1] (output1) {$y_1$};
    \node[neuron, below=of output1] (output2) {$y_2$};
    \node[neuron, below=of output2] (output3) {$\ldots$};
    \node[neuron, below=of output3] (output4) {$y_n$};
    
    \foreach \i in {1,...,4}
        \foreach \j in {1,...,4}
            \draw[->] (input\i) -- (layer\j);
    
    \foreach \i in {1,...,4}
        \foreach \j in {1,...,4}
            \draw[->] (layer\i) -- (output\j);

    \end{tikzpicture}}
\caption{A one-layer neural network with $n$ inputs and $n$ outputs, featuring a hidden layer with $n$ neurons, mathematically represented as $ \mathbf{a} = f.(\mathbf{W}_1 \mathbf{x} + \mathbf{b}_1) \rightarrow \mathbf{y} = f.(\mathbf{W}_2 \mathbf{a} + \mathbf b_2)$.}
\label{fig:multiple_neurons}
\end{figure}

%% file: sections/Background/CV.tex
\subsection{Computer Vision}\label{sec:computer_vision}
By using the structure of a neural network, images can be processed by taking their pixels as inputs. One can create an array of every pixel present in an image, this process is known as \textit{flattening}. If the picture contains \textit{color channels} (red, green, blue) these are also flattened \cite{han2021soda10m}.

If the input image has the dimensions ($1920 \times 1080 
\times 3$), where $1920$ is the width, $1080$ is the height and $3$ represents the color channels, this would yield an input layer of $1920 \times 1080 \times 3 = 6\ 220\ 800$ neurons. This is computationally costly \cite{CNN_paper}.

The \textit{Convolutional Neural Network} (CNN) provides an approach to combat this.

%% file: sections/Background/CNN.tex
\subsection{Convolutional Neural Networks}\label{sec:cnn}
The CNN is an extension of the neural network with the implementation of convolutional and pooling layers \cite{CNN_paper, CNN_paper_cracked}.

\subsubsection{Convolutional Layers} \label{sec:cnn_layer}
A convolutional layer applies a series of \textit{filters}, or \textit{kernels}, across an image, calculating the dot product between the filter weights and the input pixels at each position (see Figure~\ref{fig:conv_kernel}). These filters are designed to detect specific \textit{features} within the image, such as edges or textures \cite{CNN_paper}. These are henceforth referred to as feature maps. The (symmetrical) image consists of dimensions $(I \times I \times C)$, while the dimensions of a kernel are $(F \times F \times C)$.  A single feature map dimensions are $(O \times O \times 1)$, where $O$ is given as,

\begin{equation}
    \label{convolution_dimension}
    O = \frac{I - F + 2P}{S} + 1,
\end{equation}

where $P$ represents the padding around the border and $S$ the \textit{stride}, denoting the step size of the kernel. If there are $K$ kernels present the size of the output feature map of size $(O \times O \times K)$.

\input{imgs/latexpaint/kernel_example}

There are multiple more convolutional techniques such as \textit{depthwise convolution}\label{depthwiseConv}, the technique of using a single filter/kernel for a single channel or \textit{pointwise convolution}, where the kernel size is $1\times1$ and the stride size is $1$ \label{pointwiseConv}, effectively applying convolution pointwise \cite{liu2022convnet}.
These feature maps are then individually \textit{normalized}, put through an activation function, and \textit{pooled} \cite{aktiveringsfunktioner, CNN_paper_cracked}.

\subsubsection{Layer Normalization} 
To stabilize networks during training, normalization is performed in multiple models, including CNNs and other models. One of these normalizations are known as \textit{layer normalization} \label{sec:cnn_layer_norm} \cite{Layer_Normalization}. This type of normalization calculates the mean $\mu$ and standard deviation $\sigma$ across all activations $a_i$ from the incoming layer $l$,

\begin{equation}
    \mu^l = \frac{1}{H^l}\sum^H_{i=1}a^l_i, \quad 
    \sigma^l = \sqrt{\frac{1}{H} \sum^H_{i=1}(a_i^l-\mu^l)^2}, 
\end{equation}

where $H$ denotes the amount of hidden units in said layer. These values normalize the activations,

\begin{equation}
    y^l = \gamma^l \left( \frac{a_i^l - \mu^l}{\sqrt{\sigma^2 + \epsilon}} \right) + \beta^l,
\end{equation}

where $\gamma^l$ and $\beta^l$ are learnable parameters (optimizable as of Section \ref{sec:nn_backprop}), and $\epsilon$ is a small constant added for numerical stability ($1e-5$) \cite{PyTorchLayerNorm}.

\subsubsection{Pooling} \label{sec:cnn_pooling}
Another optimization is to downsample the dimensionality through pooling, decreasing the amount of parameters and lowering the computational cost. One example of pooling is \textit{Average pooling}, which involves computing the average value of a region of the feature map \cite{CNN_paper_cracked} with the same technique as described for the kernel in Equation \ref{convolution_dimension} and Figure \ref{fig:conv_kernel}. 

After several sequences of convolutional and pooling layers, the resulting feature representations are flattened and connected to a fully connected layer. This layer integrates the learned features to perform classification and backpropagation as described in Section \ref{Neural Network}.

%% file: imgs/latexpaint/kernel_example.tex
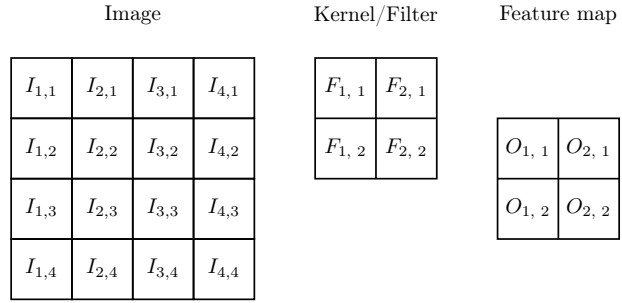
\begin{figure}[H]
\centering
    \scalebox{0.8}{
    \begin{tikzpicture}[annot/.style={text width=4cm, align=center, anchor=north}]
        \foreach \x in {0, ...,3}
            \foreach \y in {0, ..., 3}
                \draw[thick] (\x, \y) rectangle (\x+1, \y+1);
    
        \foreach \x in {2,1}
            \foreach \y in {2,1}
                \node at ( \x + 4.5, 4.5 - \y) [black] (kernel\x){$F_{\text{\x, \y}}$};
        \draw[thick] (5, 2) rectangle (7, 4); 
        \draw[thick] (6, 2) -- (6, 4); 
        \draw[thick] (5, 3) -- (7, 3); 

    \foreach \x in {4,...,1}
        \foreach \y in {4,...,1} 
                \node at (\x-0.5, 4.5-\y) {$I_{\x,\y}$};

        \foreach \x in {2,1}
            \foreach \y in {2,1}
                \node at ( \x + 7.5, 3.5 - \y) [black] (feature\x){$O_{\text{\x, \y}}$};
        \draw[thick] (8, 1) rectangle (10, 3); 
        \draw[thick] (9, 1) -- (9, 3); 
        \draw[thick] (8, 2) -- (10, 2); 

    \node[annot] at (6,5)  (Kernel)  {Kernel/Filter};
    \node[annot] at (2,5) (Image) {Image};
    \node[annot] at (9,5) (feature) {Feature map};
    \end{tikzpicture}}
    \caption{An image ($4 \times 4 \times 1$) with no padding, convolved by a $(2 \times 2 \times 1)$ kernel with 2 in stride, producing a $(2\times 2)$ feature map.}
    \label{fig:conv_kernel}   
\end{figure}

%% file: sections/Background/ConvNext.tex
\subsubsection{ConvNeXt}\label{sec:convnext}
\input{imgs/latexpaint/convnextencoder}
ConvNeXt models combine principles from Vision Transformers (see Section \ref{sec:ViT}) with traditional CNN architectures to enhance feature extraction and scalability \cite{liu2022convnet}. In this paper, the base model is used, which has been pre-trained on over one million images from the ImageNet-1K dataset \cite{liu2022convnet}.

Initially, the network employs a "patchify" mechanism  (similar to Section \ref{sec:ViT_embedding}), using $4 \times 4$, with a stride of 4, convolutions to decompose the input image. 

This is followed by several stages of ConvNeXt blocks (see Figure~\ref{fig:convnext_block}), where each block is put through the following sequence, depthwise convolutions followed by layer normalization, pointwise convolutions with four times the amount of channels from the input. The \textit{GELU} activation function (see Figure \ref{fig:activation_functions}) is then applied to these feature maps before a final pointwise convolution returns the dimensions back to the input dimensions. Some values are let through the block without interaction to keep the gradient intact, represented by the long arrow in Figure~\ref{fig:convnext_block}.

The last layer is then put through a global average pooling layer, making each feature consist of its average value \cite{liu2022convnet}. This is then connected to a multi-layered perceptron, fully connected neural network (MLP) from which predictions are made.

%% file: imgs/latexpaint/convnextencoder.tex
\begin{wrapfigure}{r}{0.15\textwidth}
\centering 
\scalebox{0.6}{
\begin{tikzpicture}[
    block/.style={rectangle, draw, text width=5em, text centered, rounded corners, minimum height=3em},
     >={Stealth[length=3mm]},
    line/.style={draw, -Latex},    
    neuron/.style={rectangle, draw, minimum size=1cm},
    layer/.style={circle, draw, minimum size=1cm},
    annot/.style={text width=4cm, align=center, anchor=north}
]
  \node [block] (depthconv) {d7$\times$7, 128};
  \node [block, below=of depthconv] (layernorm) {LN};
  \node [block, below=of layernorm] (conv1) {1$\times$1, 512};
  \node [block, below=of conv1] (gelu) {GELU};
  \node [block, below=of gelu] (conv2) {1$\times$1, 128};
  \node [circle, draw, below=of conv2, fill=white] (add) {$+$};

  \draw [->] (depthconv) -- (layernorm);
  \draw [->] (layernorm) -- (conv1);
  \draw [->] (conv1) -- (gelu);
  \draw [->] (gelu) -- (conv2);
  \draw [->] (conv2) -- (add);

  \draw [->] (depthconv.east) -- ++(2,0) |- (add.east);
  \coordinate (out) at 
  ([yshift=-1cm]add);
  \draw [->] (add) -- (out);

\end{tikzpicture}}
\caption{First ConvNeXt block for ConvNeXt\_base \cite{liu2022convnet}.}
\label{fig:convnext_block}
\end{wrapfigure}
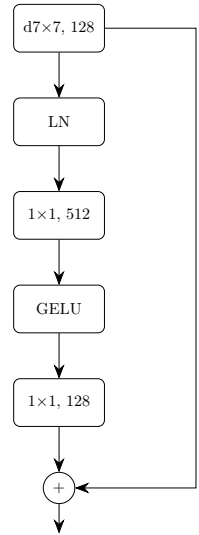




%

%% file: sections/Background/ViT.tex
\subsection{Vision Transformers} \label{sec:ViT}
Following great success in Natural Language Processing (NLP) with the usage of \textit{transformers}, the inspiration of the architecture was integrated with computer vision to create the first Vision Transformers (ViT) \cite{Visual_transformer}.
The transformer builds on the implementation of \textit{attention}, published in the 2017 paper \textit{Attention Is All You Need} \cite{vaswani2023attention}. 
\input{sections/Background/attention}
\subsubsection{Embedding} \label{sec:ViT_embedding}

Since the transformer is built on taking in a one-dimensional input, the images need to be \textit{flattened} from $\mathbf x \in \mathbb{R}^{H \times W \times C}$ into one dimension \cite{Visual_transformer}. This is achieved by transforming the picture into a sequence of flattened \textit{patches},

\begin{equation}
    \mathbf{x}_p^i \in \mathbb{R}^{1 \times (P^2 C)},\quad i = 1, \ldots, N,
\end{equation}

where $(P\times P)$ is the corresponding resolution of each patch. The whole image transformation can be expressed as $\mathbf{x}_p \in \mathbb{R}^{N \times P^2 C}$ where $N = (H \times W)/(P\times P)$. This is then linearly projected with the embedding matrix $\mathbf{E} \in \mathbb{R}^{P^2 C\times D}$ where $D$, known as the \textit{hidden space}, is the dimension in which each patch is projected onto
\begin{equation}
    \mathbf{z}_0^i = \mathbf{x}_p^i \mathbf E \in \mathbb{R}^{N\times D}.
\end{equation}

The subscript in $\mathbf z^i_0$ represents the current ViT attention block $l$ and the superscript represents the index of the patch.
This technique removes biases achieved from pooling and convolution seen in the CNN \cite{liu2022convnet}. To retain the information of the order of patches, positional embedding is added to the patches, creating the only inductive bias in the vision transformer.
A \textit{class token}\label{class_token} is inserted before the patch embedding with position $\mathbf x_p^0$, from which predictions will be made. Both positional encoding and class token are trainable parameters.

\subsubsection{Multiheaded Self-Attention} \label{sec:ViT_MSA}
After the transformation of $\mathbf x$ into $\mathbf z_0^i$ self-attention can be performed. Queries $(\mathbf{Q})$, keys $(\mathbf{K})$, and values $(\mathbf{V})$ are created by applying a linear transformation with the weight matrices $\mathbf{U}_{QKV}$

\begin{equation}
    \label{MHSA_dimensions}
    \left[\mathbf{Q}, \mathbf{K}, \mathbf{V}\right] = \mathbf{z}_0\mathbf{U}_{QKV}, \quad
    \mathbf{U}_{QKV} \in \mathbb{R}^{D \times 3D_h}.  
\end{equation}

This results in the concatenated tensor $\left[\mathbf{Q},\mathbf{K}, \mathbf{V}\right]$, from which are split into $h$ different sets of queries, keys, and values, each corresponding to a different "head" of the attention mechanism

\begin{equation}
    \mathbf{Q}_i, \mathbf{K}_i, \mathbf{V}_i \in \mathbb{R}^{N \times D_h},\quad  i = 1, \ldots, h.
\end{equation}

\input{imgs/latexpaint/ViTencoder}

The subscripts $i$ denote the $i$-th attention head, and $h$ represents the total amount of heads. Each head computes a separate attention output

\begin{equation}
    \text{Attention}(\mathbf{Q}_i, \mathbf{K}_i, \mathbf{V}_i) = \text{softmax}.\left(\frac{\mathbf{Q}_i\mathbf{K}_i^T}{\sqrt{D_h}}\right)\mathbf{V}_i.
\end{equation}

The outputs from each head are then concatenated back into one 

\begin{equation}
    \text{Concatenated Heads} = \text{Concat}(\text{Attention}_1, \ldots, \text{Attention}_h).
\end{equation}

This concatenated output is then linearly transformed back to the original hidden space $D$ using another trainable weight matrix $\mathbf{U}_{msa}$

\begin{equation}
    \text{MSA}(\mathbf{z}) = \text{Concatenated Heads} \cdot \mathbf{U}_{msa}, \quad
    \mathbf{U}_{msa} \in \mathbb{R}^{h D_h \times D}.
\end{equation}

These values are then normalized as per Section \ref{sec:cnn_layer_norm} to be connected to an MLP (multilayer perception) which contains two fully connected layers with GELU (see Figure \ref{fig:activation_functions}) and the same amount of neurons as the hidden dimension $D$. This architecture is visualized in Figure \ref{ViT block}.
This process of encoding is stacked $L$ amount of times until the final prediction is made from the class vector $\mathbf{\hat{y}} = \text{MLP}{(z_L^0)}$.

%% file: sections/Background/attention.tex
\subsubsection{Attention}\label{sec:ViT_attention}
The idea of \textit{attention} stems from creating relevancy of a certain part of the input against the entire input. This is often the case with NLPs such as Generative Pre-trained Transformers (GPTs), where given a text input, supply relevancy for the words and their order of appearance to provide context.

Neural networks can enhance their ability to focus on specific contexts by utilizing a mathematical attention model evaluating the input/embedding $\mathbf{x}$ (see Section \ref{sec:ViT_embedding}) against weight matrices.

These matrices transform the input $\mathbf{x}$ into queries ($\mathbf{Q}$), keys $(\mathbf{K})$, and values $(\mathbf{V})$ (see \ref{MHSA_dimensions}). 
The prediction for the input $\mathbf{x}$ is computed as,

\begin{equation}
    \text{Attention}(\mathbf{Q}, \mathbf{K}, \mathbf{V}) = \text{softmax}.\left(\frac{\mathbf{Q}\mathbf{K}^T}{\sqrt{d_k}}\right)  \mathbf{V},
\end{equation} 

where the dot (.) once again indicates that the function acts on each element of the vector. 

These matrices are trained to extract features from data points with the underlying idea of backpropagation motivated in Section \ref{sec:nn_backprop}. Attention can similarly be used in vision transformers where attention is performed to determine possible points of relevancy to a classifier/class token \cite{Visual_transformer, vaswani2023attention}.

\begin{figure}[H]
    \centering
    \includegraphics[width=0.4\textwidth]{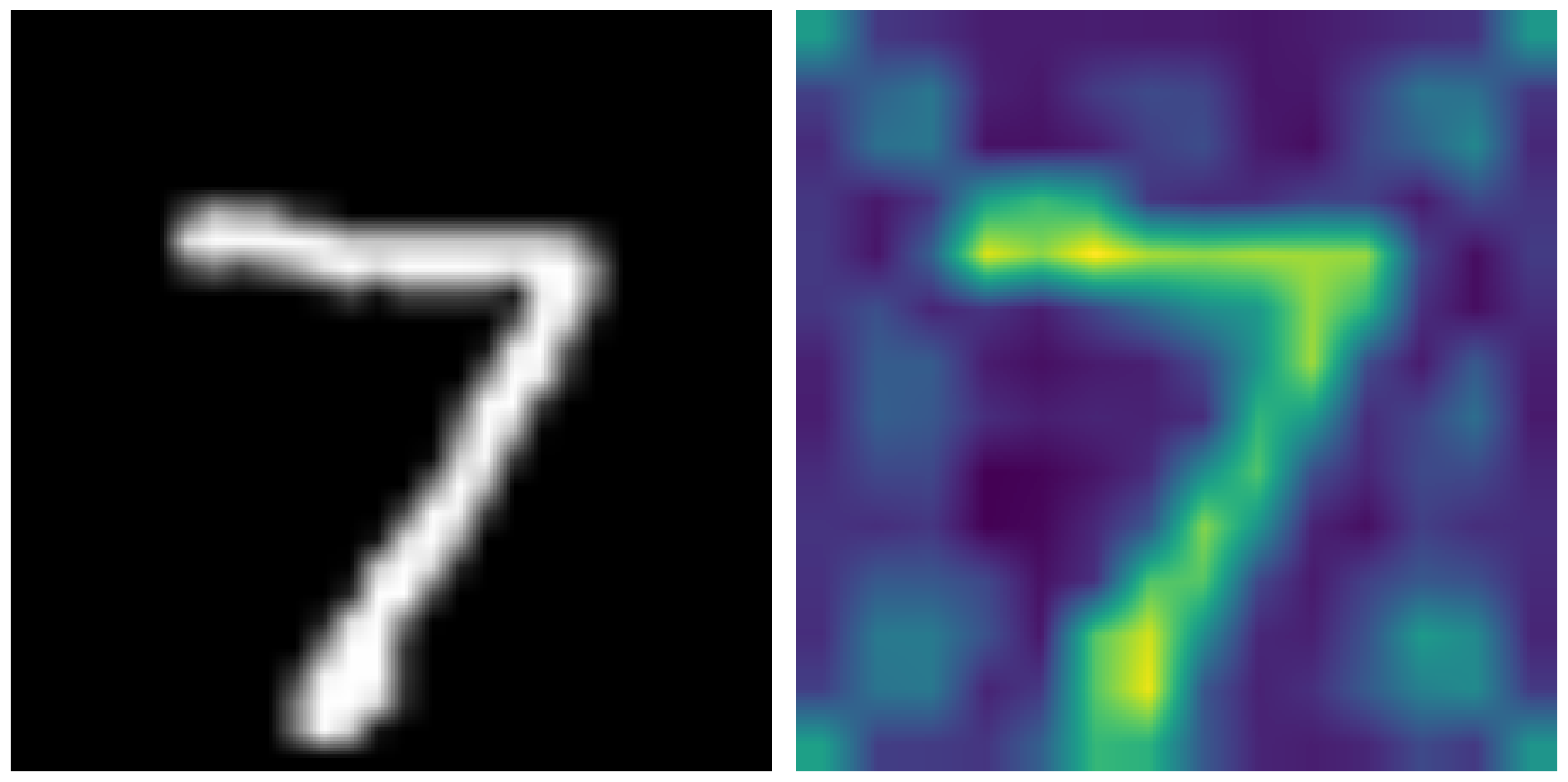}
    \caption{Visualization of the attention mechanism.}
    \label{fig:attention_viz}
\end{figure}

%% file: imgs/latexpaint/ViTencoder.tex
\begin{wrapfigure}{r}{0.175\textwidth}
\centering
\scalebox{0.6}{
\begin{tikzpicture}[
    node distance=2cm,
    line width=1pt,
    >={Stealth[length=3mm]},
    boxes/.style={rectangle, thick, fit=#1, fill=gray!5},
    neuron/.style={circle, draw, minimum size=1cm, fill=white},
    process/.style={rectangle, minimum width=3cm, minimum height=1cm, text centered, draw=black, fill=white}]

    \node (in) [process] {Embedded Patches};
    \node (norm1) [process, below of=in] {LN};
    \node (mha) [process, below of=norm1] {MSA};
    \node (add1) [neuron, below of=mha] {$+$};
    \node (norm2) [process, below of=add1] {LN};
    \node (mlp) [process, below of=norm2] {2 MLP, GELU};
    \node (add2) [neuron, below of=mlp] {+};
   
    \begin{scope}[on background layer]
        \node[boxes=(norm1) (mha) (add1) (norm2) (mlp) (add2), inner ysep=20pt,  yshift=1mm, inner xsep=22pt, draw, thick] (custom_box) {};
    \end{scope}
    \node [anchor=south west] at (custom_box.south west) {$L\times$};

    \draw [->] (in) -- (norm1);
    \draw [->] (norm1) -- (mha);
    \draw [->] (mha) -- (add1);
    \draw [->] (add1) -- (norm2);
    \draw [->] (norm2) -- (mlp);
    \draw [->] (mlp) -- (add2);
    \coordinate (out) at 
    ([yshift=-1.75cm]add2);
    \draw [->] (add2) -- (out);

    \coordinate (midpoint1) at ($(in)!0.45!(norm1)$);
    \draw [->] (midpoint1) -| ([xshift=-0.4cm]norm1.west) |- (add1);
    \coordinate (midpoint2) at ($(add1)!0.45!(norm2)$);
    \draw [->] (midpoint2) -|([xshift=-0.4cm]norm2.west)|- (add2);

\end{tikzpicture}}
\caption{vision transformer encoder/block \cite{Visual_transformer}.}\label{ViT block}
\end{wrapfigure}
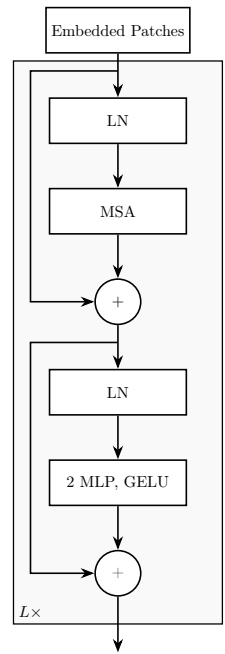

%% file: sections/Background/UQ.tex
\subsection{Uncertainty in Information Theory} \label{sec:uncertainty}

Let $X$ be the discrete random variable that takes values in the event space $\chi$.

In information theory, the Shannon entropy $H$ of $X$ is defined as  \cite{shannon1948mathematical},

\begin{equation}
    \label{shannon_entropy}
    H(X) = -\sum_{x \in \chi} p(x)\ln p(x),
\end{equation}

where $p(x) = P(X = x)$. This can be condensed to the \textit{expected uncertainty}. We can rewrite the equation as

\begin{equation}
    \label{shannon_entropy_ex}
    H(X) = \EX[-\ln p(X)].
\end{equation} 

%% file: sections/Method/hypothesis.tex
\subsection{Entropy-based Uncertainty Quantification Framework} \label{sec:entropy_framework}
This section explains the proposed framework motivated by Shannon entropy, see Equation \ref{shannon_entropy}.

\subsubsection{Perturbation in Neural Networks} \label{sec:entropy_pert_nn}
Upon completion of training, a model yields a final weight matrix, $\mathbf{W}$ with all the trainable weights of a network (in reality this is a weight \textit{tensor} where each element is a weight matrix for each layer, but the same theory and logic can be applied to a matrix). If the model comprehensively understood the problem beyond the specific training dataset we define the weight matrix as the \textit{ideal weight matrix}. Conversely, a failure to identify these ideal weights will result in sub-optimal performance on similar unseen datasets.

To assess whether a model has successfully acquired ideal weights, we hypothesize one can introduce perturbation to the weight matrix $\mathbf{W}$ and observe the impact on performance. Here, the perturbation magnitude is scaled by a scalar $\sigma$, where a higher $\sigma$ correlates to greater disturbance. A notable performance dip at small $\sigma$ levels should suggest that the model is not confident and it is hypothesized that a small amount of perturbation should not offset the prediction.

\subsubsection{Distribution of Perturbed Matrices} \label{sec:entropy_dist_pert}
For a given input image $\mathbf{x}$ and weight matrix $\mathbf{W}$, the network produces a probability vector $\hat{\mathbf{y}} = \hat{\mathbf{y}}(\mathbf{x}, \mathbf{W})$.

From this, a point estimate can be obtained by applying the \textit{argmax} defined as,

\begin{equation}
    \label{eq:argmax}
    \hat{y}(\mathbf{x}) = \textbf{argmax}(\mathbf{\hat{y}}(\mathbf{x})),
\end{equation} 

where the argmax function yields the class index of the element with the highest probability in the probability vector $\mathbf{\hat{y}}$.

Our method is in general applicable to predictors that give point estimates. 

We use
\begin{equation}
    \hat{y} = \mathcal{F}(\mathbf{x}, \mathbf{W}),
\end{equation}

to denote the functional dependency of the predicted output $\hat{y}$ on the weight matrix $\mathbf{W}$ and the input image $\mathbf{x}$ or more generally an estimator taking in an input $\mathbf {x}$ and a high-dimensional weight vector $\mathbf{W}$ to produce a point estimate $\hat{y}$.

We propose to perturb the weight matrix $\mathbf{W}$ with Gaussian noise,

\begin{equation}
    \mathbf{W}_\sigma  = \mathbf{W} + \sigma \mathbf{N},
\end{equation}

where $\mathbf{N} \sim \mathcal{N}(0, 1)$ symbolizes a Gaussian noise matrix, with the same shape of $\mathbf{W}$, and $\sigma$ the scalar impact of the matrix $\mathbf{N}$. With this motivation, we induce randomness in the network predictions,

\begin{equation}
  {\hat{y}}_{\sigma}(\mathbf{x})  = \mathcal{F}(\mathbf{x}, \mathbf{W}_\sigma),
\end{equation}

where we consider the input image $\mathbf{x}$, applied to the network, with the perturbed weight matrix $\mathbf{W}_{\sigma}$. This yields the argmax prediction ${\hat{y}}_\sigma$ of the class of $\mathbf{x}$.

It is important to note that perturbing the weights and creating a single prediction constitutes a random experiment. It is therefore meaningful to examine the probability distribution of the random variable $\mathcal{F}(\mathbf{x}, \mathbf{W}_\sigma)$ for a fixed input and random weights, or for example its entropy $H_\sigma(\mathbf{x})$ (which both depend on the input $\mathbf{x}$).

By repeating the experiment for a given input $\mathbf{x}$, creating samples of $\mathcal{F}(\mathbf{x}, \mathbf{W}_{\sigma})$ but drawing different samples $\mathbf{W}_{\sigma}^{(i)}$, $i = 1, \dots, N$, obtaining potentially different prediction $\hat{y}^{(i)}$  we can empirically investigate this.

With this motivation, we search for the underlying distribution of the model and the properties of said distribution.

\subsubsection{PI: Perturbation Index} \label{sec:entropy_pi}
For image classification tasks, accuracy stands as an important metric. The accuracy is defined as the fraction of correct predictions based on the argmax-prediction (Equation \ref{eq:argmax}) the model gets. To quantify the impact of perturbation on model accuracy, we introduce the Perturbation Index (PI) defined by

\begin{equation}
    \pi_{\sigma} = \alpha - \alpha_{\sigma},
\end{equation}

where $\alpha$ represents the original model's accuracy, and $\alpha_{\sigma}$ denotes the \emph{expected} accuracy post perturbation for a given sample. Note that these accuracies are calculated on the \textbf{entire} dataset $\mathcal{D}$ averaging over all draws of $\mathbf{W}_\sigma$.

This measures how far the perturbed model has deviated from the original model and its accuracy. Therefore, given a well-trained model, low PI metric value indicates a robust model.

\subsubsection{PSI: Perturbation Stability Index} \label{sec:entropy_psi}
To assert the uncertainty of the prediction of the model, we define the Perturbation Stability Index (PSI). 

We propose that this index suggests the prediction's inherent classification uncertainty. For a given input, if the model generates varying argmax-predictions, under perturbation, it suggests uncertainty of the classification. In other words, there should be a negative correlation between prediction accuracy and the Shannon entropy of these predictions.

By quantitatively estimating this correlation through sampling, we can assess uncertainty for a single input without seeing the ground truth. A lack of deviation from the model's predictions indicates no error. Essentially, the model's accuracy is linked to its entropy under a specific perturbation level.

The entropy of the argmax-prediction for the perturbed output given an input $\mathbf{x}$ is defined as Shannon entropy (Equation \eqref{shannon_entropy}) of the random  variable $\mathcal F(\mathbf{x}, \mathbf{W}_\sigma)$ and can be computed from repeated samples $\hat{y}^{(i)}$, $i =1 \dots n$ as

\begin{equation}
    H_{\sigma}(\mathbf x) = \lim_{n \to \infty} - \sum_{c \in C} p^{(n)}_c \ln\left(p^{(n)}_c\right),
\end{equation}

where $p^{(n)}_c$ is the proportion of predictions $\hat{y}^{(i)}$ equal to the  class index $c$ out of all $C$ classes in the $n$ samples of $\hat{y}$ for given $\mathbf{x}$. 

We define 

\begin{equation}
\beta_{\sigma}(X, Y) = \begin{cases}
    1 &  \hat{y}_{\sigma}(X) = Y\\
    0 & \text{else},
    \end{cases}
\end{equation}

as the correctness of the prediction for a random pair of an image and the corresponding class label from $(X, Y) \sim \mathcal{D}$, where the randomness of $\beta$ stems from drawing random samples from $\mathcal{D}$. Note

\begin{equation}
    \mathbb E [\beta_{\sigma}(X, Y)] = \alpha_{\sigma},
\end{equation}

is the overall accuracy value for a given perturbation $\sigma$.

Finally, the function mapping $\mathbf x$ to 
\begin{equation}
    \label{eq:p_sigma}
    p_\sigma(\mathbf x) = P\left(\hat y_\sigma(X) = Y \big| H_\sigma(X) = h\right)
    \quad \text{ where }\quad h = H(\mathbf x),
\end{equation}

allows us to look at the data and compute the accuracy of the prediction the noised model would make without having seen the corresponding correct label.

$p_{\sigma}(\mathbf{x})$ can be understood as the probability of making a correct prediction within all draws from the data which have \emph{the same entropy as $\mathbf{x}$}.
 
Here, empirically determining the function $\mathbf{x} \mapsto p_{\sigma}( \mathbf x)$ allows us to calibrate the nonlinear relationship between the entropy of (possibly new) inputs $\mathbf x$ and accuracy of the network $\mathcal F$. 

To choose a good value of $\sigma > 0$ we propose to consider the inverse correlation between uncertainty (entropy) and accuracy as objective,

\begin{equation}
    \psi_{\sigma} = \alpha_\sigma - \text{corr}(\beta_{\sigma}(X, Y), H_{\sigma}(X)) \cdot \lambda,
\end{equation}

where again the randomness comes from $H_\sigma(X)$, $(X, Y) \sim \mathcal D$. The entropy of each input $\mathbf x$ is deterministic, but $H_\sigma(X)$ denotes the entropy of a random sample from the data distribution. 

$\lambda$ adjusts the importance of correlation on the PSI metric, therefore, a higher PSI value indicates a more stable model. Note that you can only compare different PSI values for the same $\lambda$ since it shifts the importance of the correlation.

%% file: sections/Method/dataset.tex
\subsection{Technical Choices and Dataset} \label{sec:tech_choices_dataset}
In this project the PyTorch library was used to produce neural network models \cite{Pytorch}. The PyTorch 
library is widely accredited, used in academic work, and has a large space in the implementations of neural networks \cite{NVIDIAPyTorch2024}. The models used were trained and evaluated on the MNIST dataset, consisting of $70\:000$ hand-drawn single digits with 10 labels \cite{mnist_dataset}. $60\:000$ are used for training and $10\:000$ for evaluating \& testing the model.


%% file: sections/Method/pertubation_injection.tex
\subsection{Perturbation Injection} \label{sec:pert_injection}
As described in Section \ref{sec:entropy_dist_pert} perturbation should affect the output of a model. Thus, where the perturbation is introduced in the model is of utmost interest. If the perturbation is exclusively introduced at specific layers/parameters or the whole network is perturbed, different results are hypothesized to occur. Therefore, the introduction of perturbation on the following levels will be studied:

\begin{itemize}
    \item The input image
    \item All the parameters/weights of a network
\end{itemize}

Note that for some high values of $\sigma$ for example $10.0$, the perturbed image is arbitrary. In other words, the underlying information of the original image is no longer present in the input to the network. This is because after adding noise, the color values are clamped between $0$ and $1$.

%% file: sections/Method/training_implementation.tex
\subsection{Training and Implementation of Models} \label{sec:training_impl}
As motivated in the Section \ref{sec:scope} multiple different models are tested and explained below. The same baseline has been used to train all models:

\begin{itemize}
    \item The Adam optimizer \cite{AdamOptimizer} was used for all three models, all using the same parameters for the optimizer.
    \item The average cross-entropy loss function, as mentioned in Section \ref{sec:nn_train_loss}, was used for all models.
    \item All models have been trained for 10 epochs - meaning that the learning algorithm goes through the dataset 10 times, as noted in Section \ref{sec:nn_backprop}.
\end{itemize}

This approach is chosen as a baseline for all models as a hypothetical fair comparison.

%% file: sections/Method/Models/naive.tex
\subsubsection{Na\"ive Multinomial Regression} \label{sec:impl_naive}
The Na\"ive multinomial Regression model builds on the foundation motivated in Section \ref{sec:computer_vision}, a neural network consisting of as many neurons as there are (in this case) pixels as the input. For the MNIST dataset, this would reflect $28\times28 = 784$ neurons. This is expressed mathematically as (no bias $\mathbf{b}$)

\begin{equation}
    \mathbf{y} = \mathbf{W}\mathbf{x} \; \; \; \mathbf{W} \in \mathbb{R}^{10\times 784}, \mathbf{x}\in\mathbb{R}^{784  \times 1}  \implies \mathbf{y}\in\mathbb{R}^{10\times1}.
\end{equation}

\input{imgs/latexpaint/logistic}

Since the na\"ive multinomial model (Figure \ref{fig:naive_arch}) is the most simple model, perturbation injection is only possible on two levels; At the input layer of the network and all the weights of the network.

%% file: imgs/latexpaint/logistic.tex
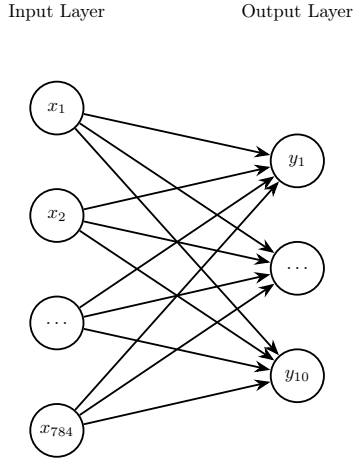
\begin{figure}[H]
\centering
\scalebox{0.7}{
\begin{tikzpicture}[
    node distance=1cm,
    line width=1pt,
    >={Stealth[length=3mm]},
    neuron/.style={circle, draw, minimum size=1cm},
    layer/.style={rectangle, draw, minimum height=1cm, minimum width=2.5cm,},
    annot/.style={text width=4cm, align=center, anchor=north}
]

\node[annot, above=of input1]  (input layer)  {Input Layer};
\node[annot,  right=0.25cm of input layer] (output layer)  {Output Layer};

\node[neuron]                   (input1) {$x_1$};
\node[neuron, below=of input1]  (input2) {$x_2$};
\node[neuron, below=of input2]  (input3) {$\ldots$};
\node[neuron, below=of input3]  (input4) {$x_{784}$};

\node[neuron, below=2.0cm of output layer] (output1) {$y_1$};
\node[neuron, below=of output1] (output2) {$\ldots$};
\node[neuron, below=of output2] (output3) {$y_{10}$};

\draw[->] (input1) -- (output1);
\draw[->] (input1) -- (output2);
\draw[->] (input1) -- (output3);

\draw[->] (input2) -- (output1);
\draw[->] (input2) -- (output2);
\draw[->] (input2) -- (output3);

\draw[->] (input3) -- (output1);
\draw[->] (input3) -- (output2);
\draw[->] (input3) -- (output3);

\draw[->] (input4) -- (output1);
\draw[->] (input4) -- (output2);
\draw[->] (input4) -- (output3);

\end{tikzpicture}} 
\caption{The Na\"ive multinomial Regression Model Architecture}
\label{fig:naive_arch}
\end{figure}

%% file: sections/Method/Models/ConvNext.tex
\subsubsection{ConvNeXt Model} \label{sec:impl_convnext}
The ConvNeXt architecture consists of a sequence of ConvNeXt blocks (\ref{fig:convnext_block}). The overall structure of the process is $C = (128, 256, 512, 1024), \; B = (3, 3, 27, 3)$, where every $B$ represents the amount of block-sequences in that specific part and $C$ the amount of channels for that specific sequence. For example, the \ref{fig:convnext_block} is the first block in that block-sequence, which has a channel of $128$.

Since the network needs three color channels as an input and the MNIST dataset only consists of one, the dataset is transformed through grey scaling to fit the channels. A similar argument is made for the size of the image, this is solved by resizing the dataset to the correct dimensions.

\begin{figure}[H]
    \centering
    \includegraphics[width=0.73\textwidth]{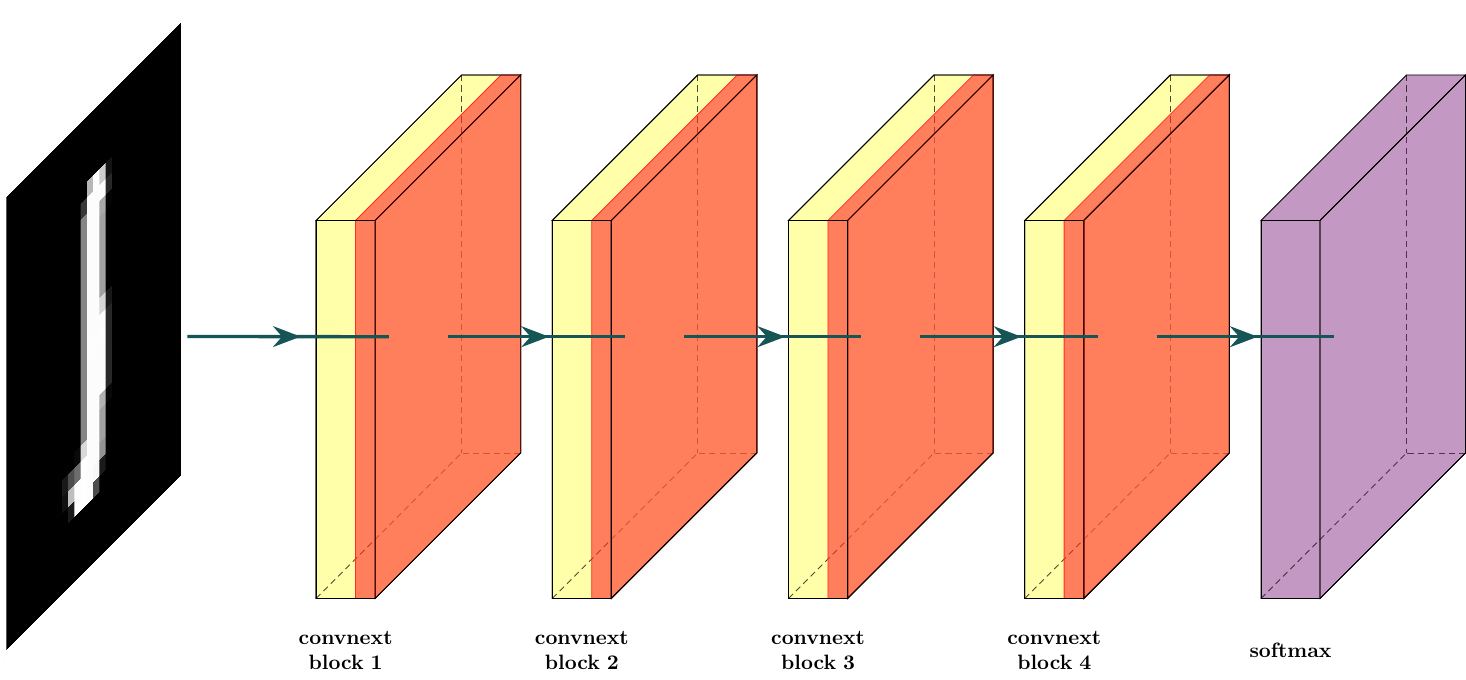}
    \caption{The High Level Architecture for the Base ConvNeXt Model}
    \label{fig:convnext_arch}
    
\end{figure}

By connecting the last layer to a fully connected linear layer (see Figure \ref{fig:convnext_arch}) with the same amount of prediction neurons as classes, all pre-trained weights gradients except for this layer can be frozen and the model can now be trained for the correct dataset.

%% file: sections/Method/Models/ViT.tex
\subsubsection{Vision Transformer Model} \label{sec:impl_ViT}
Staying true to the encoder architecture referenced in Figure \ref{ViT block}, the model used in our work was the \textit{base} model, providing a hidden size $D$ of 768, 12 encoder blocks (see Figure \ref{ViT block}) each containing 12 heads and an MLP size of 3072.
\input{imgs/latexpaint/ViT_arch}


%% file: imgs/latexpaint/ViT_arch.tex
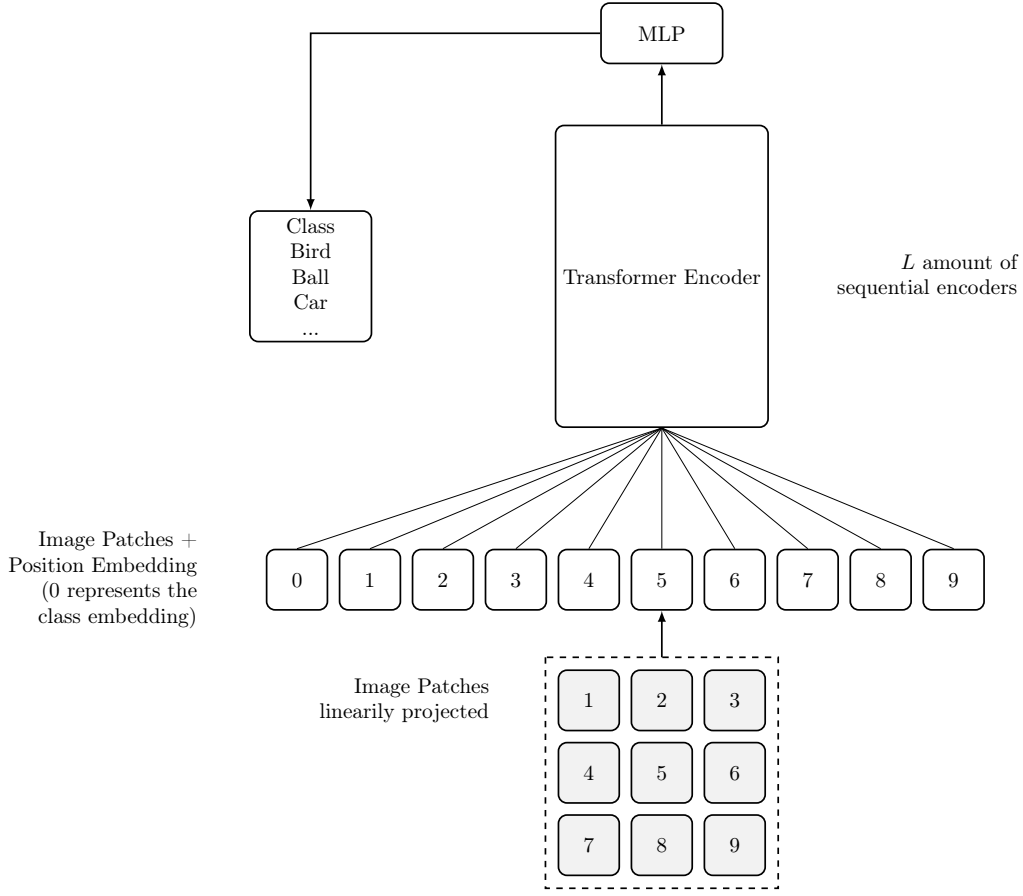
\begin{figure}[H]
    \centering
    \scalebox{0.8}{
    \begin{tikzpicture}[
        node distance=1cm,
        patch/.style={draw, thick, rectangle, rounded corners, minimum size=1cm},
        mlp/.style={draw, thick, rectangle, rounded corners, minimum height=1cm, minimum width=2cm, align=center},
        encoder/.style={draw, thick, rectangle, rounded corners, minimum height=5cm, minimum width=2cm, align=center},
        pic_patch/.style={draw, thick, rectangle, rounded corners, minimum size=1cm, fill=gray!10},
        label/.style={text width=1cm, align=center}
    ]
    
    \node (encoder) [encoder] {Transformer Encoder};
    
    \node (mlp) [mlp, above=of encoder] {MLP};
    \draw[-latex,thick] (encoder) -- (mlp); 
    
    \foreach \x in {0,...,2}
        \foreach \y in {0,...,2} {
            \pgfmathtruncatemacro{\patchnum}{\x + 3*\y + 1}
            \node (p\patchnum) [pic_patch, below=4cm of encoder.south, xshift=\x*1.2cm - 1.2cm, yshift=-\y*1.2cm] {\patchnum}; 
        }
    
    \draw [thick, dashed] ($(p1.north west)+(-0.2,0.2)$) rectangle ($(p9.south east)+(0.2,-0.2)$);
    
    \node [left=of p1.west, anchor=east, align=right] {Image Patches \\linearily projected};
    
    \foreach \i in {-1,...,8} {
        \pgfmathtruncatemacro{\patchnum}{\i + 1}
        \node (box\patchnum) [patch, below=2cm of encoder.south, xshift=\i*1.2cm - 4.8cm] {\patchnum};
        \draw[-] (box\patchnum.north) -- (encoder.south); 
    }

    \draw[-latex,thick] ($(p2.north) + (0,0.2)$) -- ($(box5.south) - (0,0)$);

    \node [right=of encoder.east, align=right] {$L$ amount of \\sequential encoders};

    \node [left=of box0.west, anchor=east, align=right] {Image Patches +\\Position Embedding\\ (0 represents the \\ class embedding)};
    
    \node (class) [mlp, left=3cm of encoder] {Class\\Bird\\Ball\\Car\\...};
    \draw[-latex,thick] (mlp) -| (class);
    
    \end{tikzpicture}
    }
    \caption{Vision Transformer Architecture}
    \label{fig:vit_architecture}
\end{figure}

%% file: sections/Results/results.tex
The results from evaluating each model listed in Section \ref{sec:scope} used the following common arguments:

\begin{itemize}
    \item $n = 10$, for the amount of iterations when calculating the different metrics.
    \item $\sigma = 0.1, 0.5, 1.0, 10.0$
    \item $\lambda = 0.1, 0.5, 1.0, 2.0$
\end{itemize}

The entropy-accuracy-certainty (EAC) graphs are produced by calculating the entropy and accuracy as described in Section \ref{sec:entropy_psi}. 

Certainty is the average probability of the correct class from the sampled outputs, defined as 

\begin{equation}
    c = \frac{1}{n} \sum_{i = 1}^{n} \hat{y}_{\sigma, Y},
\end{equation}

where $\hat{y}_{\sigma, Y}$ is the probability of the \textit{correct} class label the network outputted (note that this is \textbf{not} the argmax output).

The filled-in crosses are the expected accuracies for entropy windows defined as,

\begin{equation}
    p_{\sigma} = \lim_{\epsilon \to 0} P\left(\hat y_\sigma(X) = Y \big| H_\sigma(X) \in [h - \epsilon, h + \epsilon] \right)
\end{equation}

similarly to Equation \eqref{eq:p_sigma}, note that $h$ is not fixed, but an entropy within an interval (with the motivation being that in practice no inputs will have the \textit{exact} same entropy, so we ''relax'' the condition to be within an interval). This value is computed for multiple sub-intervals of all samples. The dashed line is the regression line of all these computed $p_{\sigma}$ points. So we expect a negative slope for the regression line as discussed in Section \ref{sec:entropy_psi}.

The EAC bar plot is produced by counting the average entropy, accuracy, and certainty for each individual class and then dividing the sums by the total count of each class. Since entropy is a value between $0$ and $\ln(\text{classes})$ it is represented as the percentage of maximum entropy in the bar plot where classes are 10 in this case.

\begin{figure}[H]
    \centering
    \begin{subfigure}[b]{0.4\textwidth}
        \includegraphics[width=\textwidth]{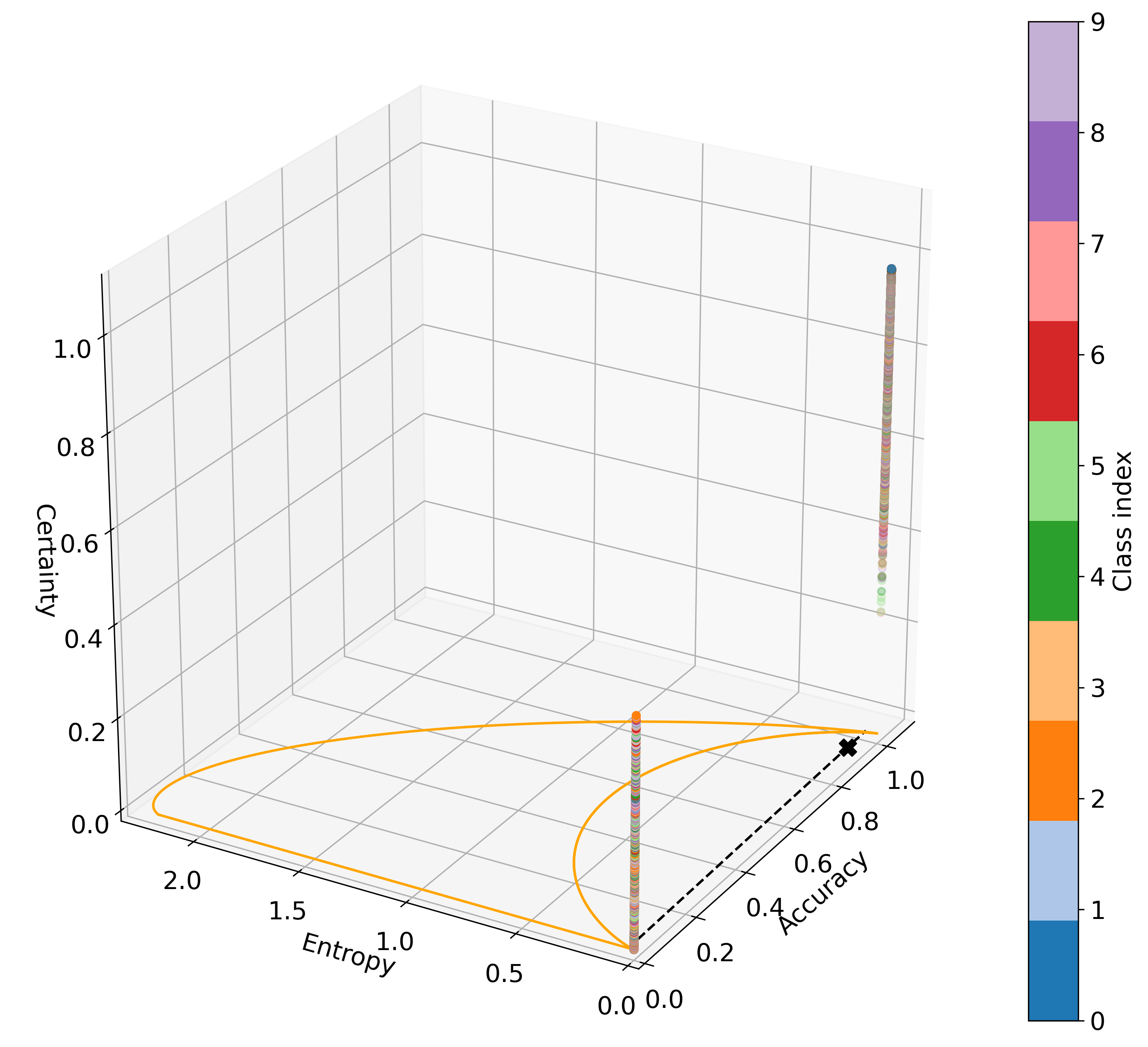}
        \caption{Example of evaluation of a model without any perturbation}
        \label{fig:example_no_perturbation_eac}
    \end{subfigure}
    \hfill
    \begin{subfigure}[b]{0.5\textwidth}
        \includegraphics[width=\textwidth]{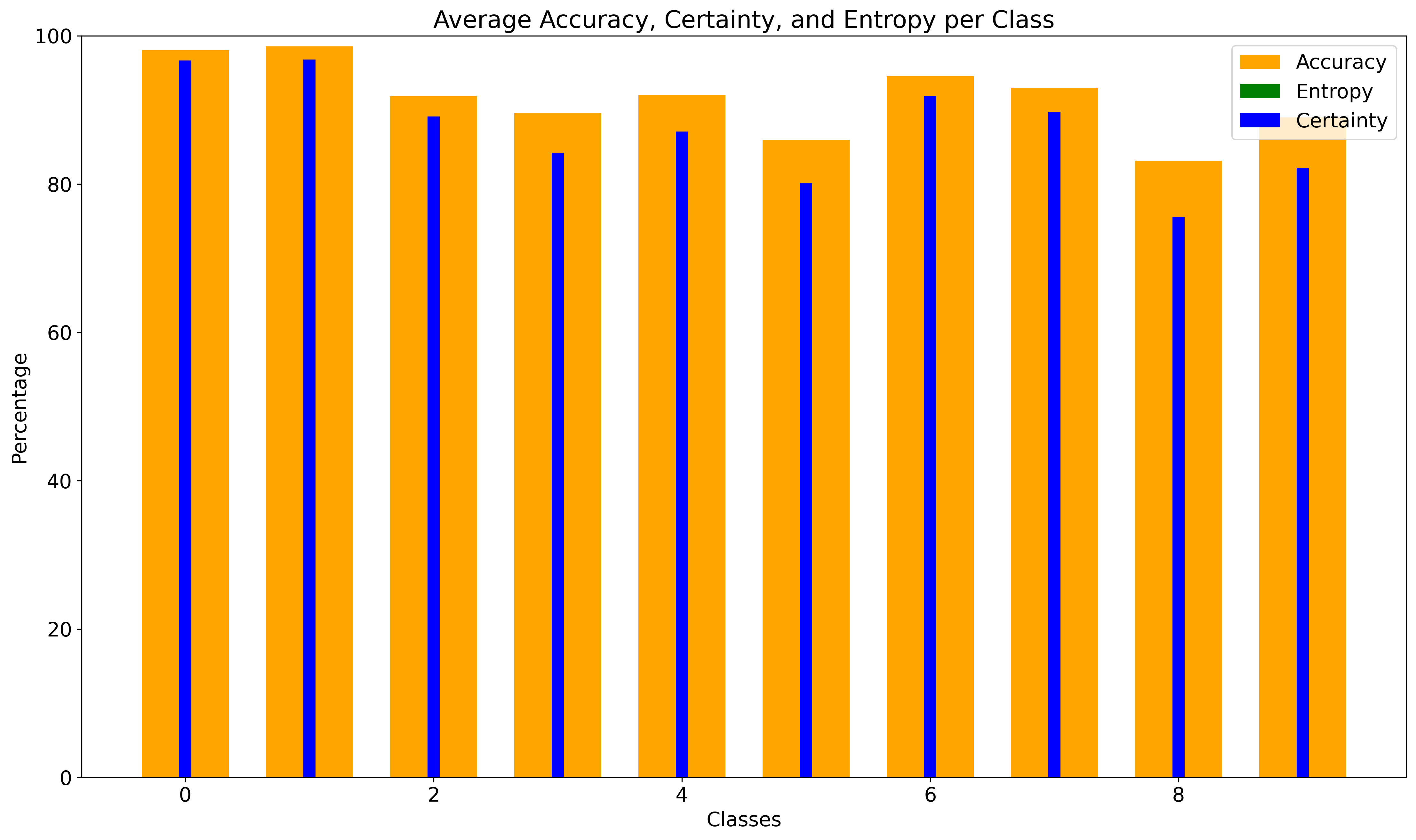}
        \caption{Example of average entropy, accuracy, and certainty on a model without perturbation}
        \label{fig:example_no_perturbation_barplot}
    \end{subfigure}
    \caption{Example of EAC graph and EAC bar plot with no perturbation}
\end{figure}

If no perturbation is introduced within a system, no entropy is present, the same input will always yield the same output. Therefore, the accuracy will always be $100\%$ or $0\%$ for any given sample in this case (see Figure~\ref{fig:example_no_perturbation_eac} and Figure~\ref{fig:example_no_perturbation_barplot} for examples).

\subsection{Na\"ive Multinomial Regression}
\begin{table}[H]
    \caption{PSI and PI Metrics for Na\"ive multinomial Regression Model}
    \centering
    \begin{tabular}{| c | c c c c | c c c c | c | c |}
        \hline
        \multirow{2}{*}{\diagbox{$\sigma$}{$\lambda$}} & \multicolumn{4}{c|}{$\psi_w$} & \multicolumn{4}{c|}{$\psi_i$} & \multirow{2}{*}{$\pi_w$} & \multirow{2}{*}{$\pi_i$} \\
        & 0.1 & 0.5 & 1 & 2 & 0.1 & 0.5 & 1 & 2 & & \\
        \hline
        0.1 & 0.965 & 1.226 & 1.551 & 2.203 & 0.939 & 1.122 & 1.351 & 1.808 & 0.021 & 0.024 \\
        0.5 & 0.679 & 1.025 & 1.458 & 2.324 & 0.345 & 0.513 & 0.722 & 1.142 & 0.310 & 0.614 \\
        1   & 0.423 & 0.712 & 1.072 & 1.793 & 0.200 & 0.295 & 0.412 & 0.648 & 0.588 & 0.741 \\
        10  & 0.125 & 0.157 & 0.197 & 0.277 & 0.099 & 0.108 & 0.119 & 0.140 & 0.783 & 0.819 \\ 
        \hline
    \end{tabular}
    \label{tab:naive_logistic_regression}
\end{table}

Looking at Table~\ref{tab:naive_logistic_regression} we can see a correlation between worse metrics with higher perturbation. It is also evident that the model performs worse with image perturbation compared to the weight perturbation on the PSI metric.

\begin{figure}[H]
    \centering
    \begin{subfigure}[b]{0.4\textwidth}
        \centering
        \includegraphics[width=\textwidth]{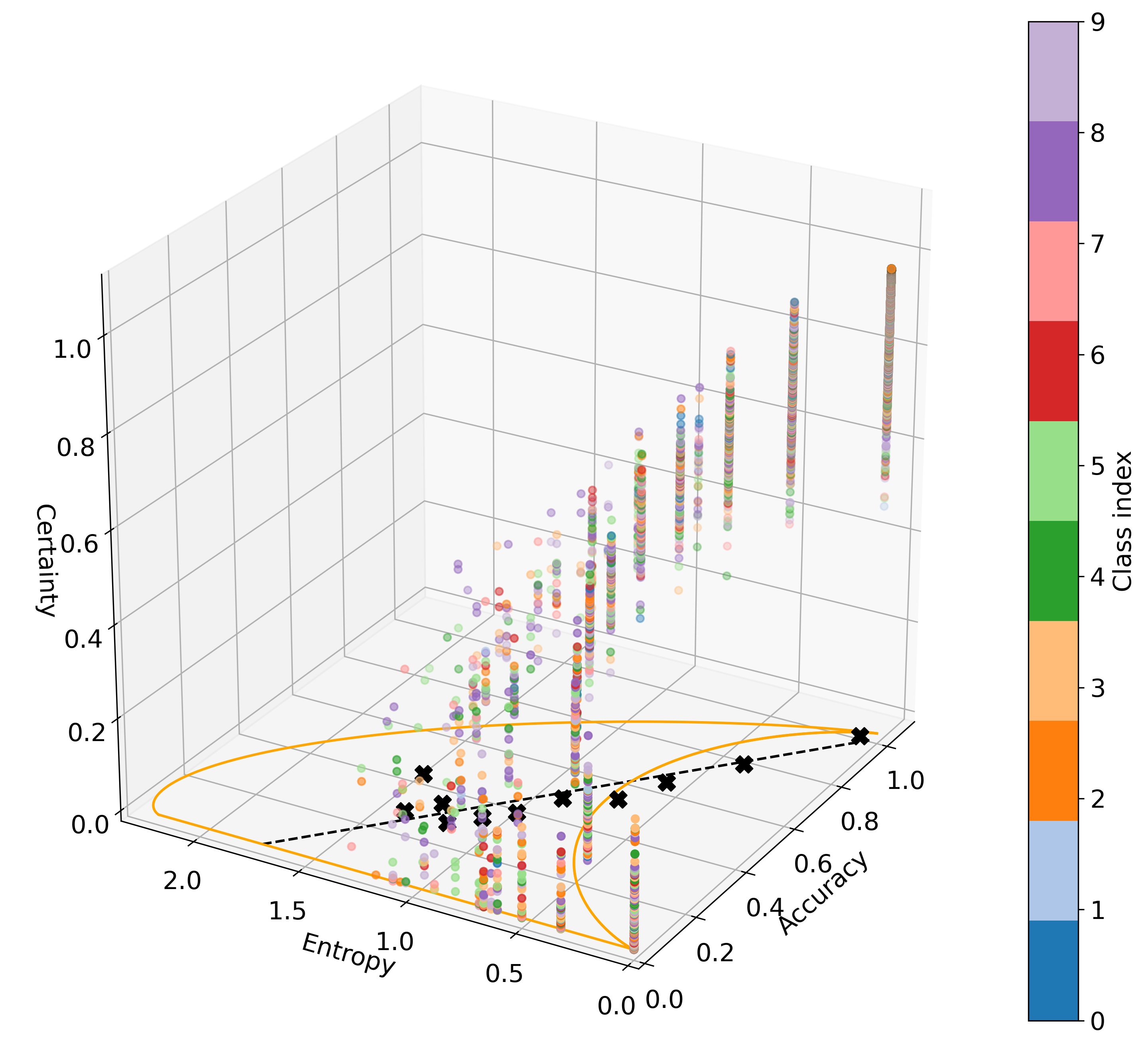}
        \caption{Weight perturbation with $\sigma = 0.1$}
        \label{fig:weight_logistic_0.1}
    \end{subfigure}
    \hfill
    \begin{subfigure}[b]{0.4\textwidth}
        \centering
        \includegraphics[width=\textwidth]{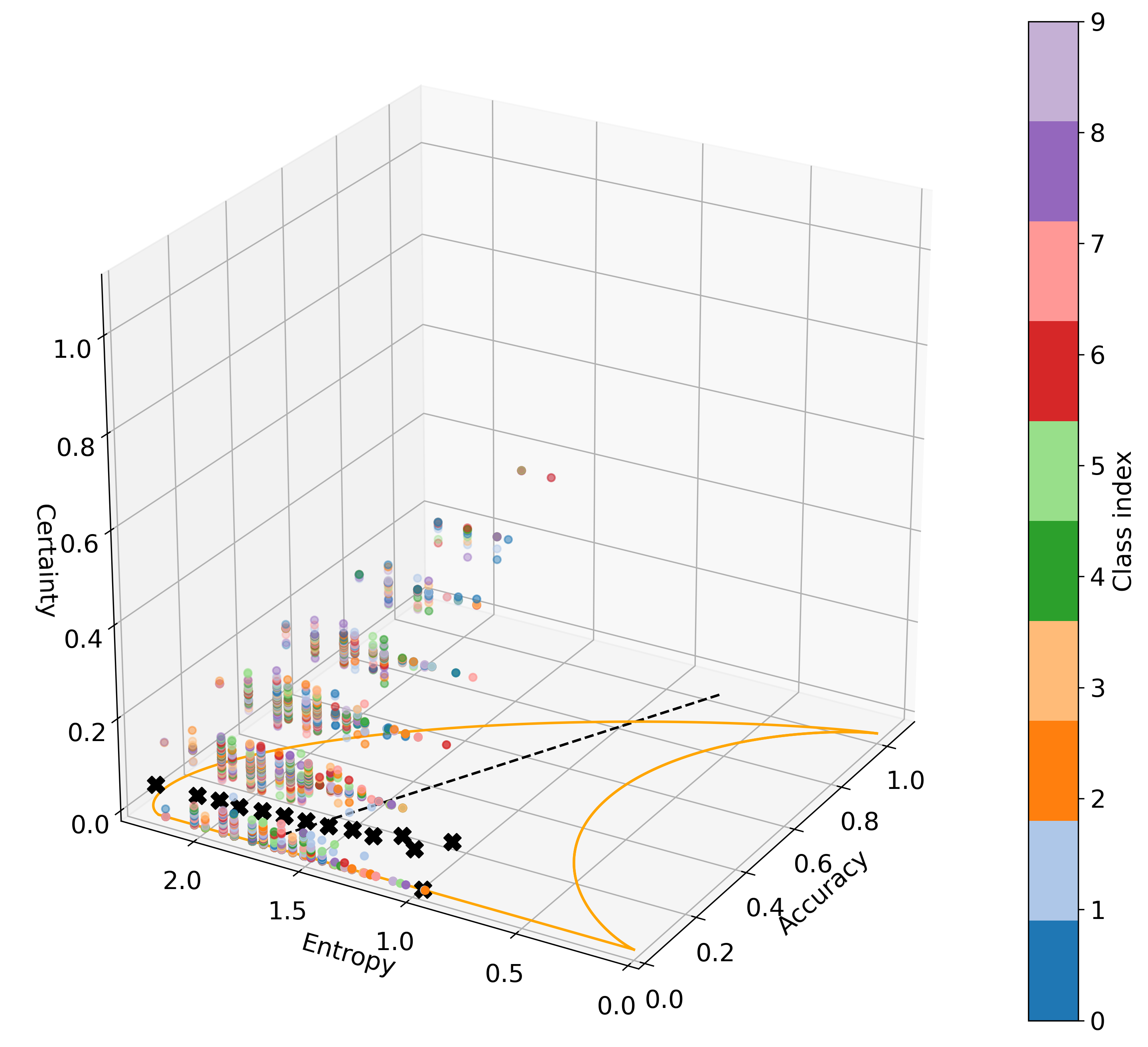}
        \caption{Weight perturbation with $\sigma = 10.0$}
        \label{fig:weight_logistic_10}
    \end{subfigure}

    \begin{subfigure}[b]{0.4\textwidth}
        \centering
        \includegraphics[width=\textwidth]{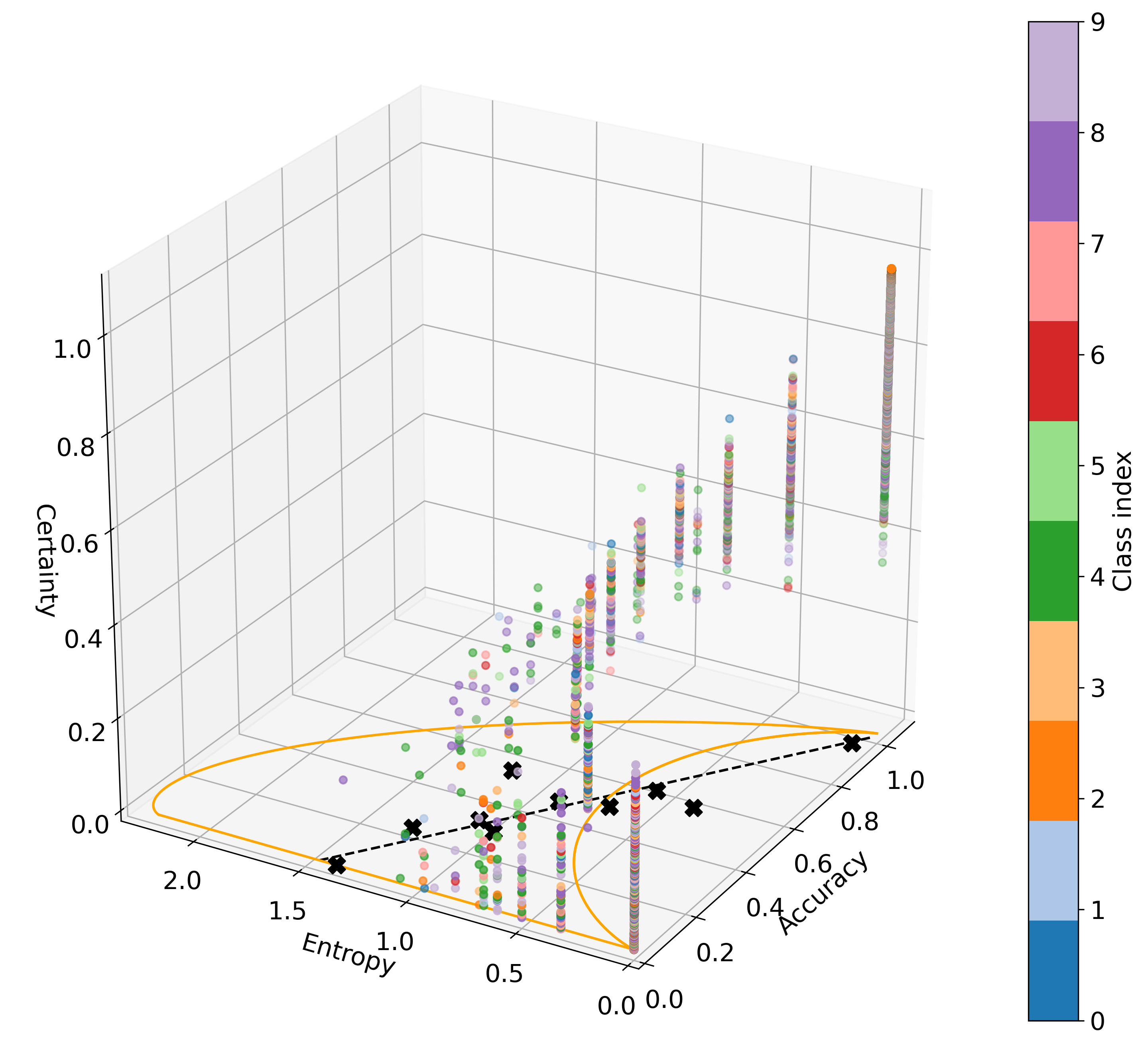}
        \caption{Image perturbation with $\sigma = 0.1$}
        \label{fig:image_logistic_0.1}
    \end{subfigure}
    \hfill
    \begin{subfigure}[b]{0.4\textwidth}
        \centering
        \includegraphics[width=\textwidth]{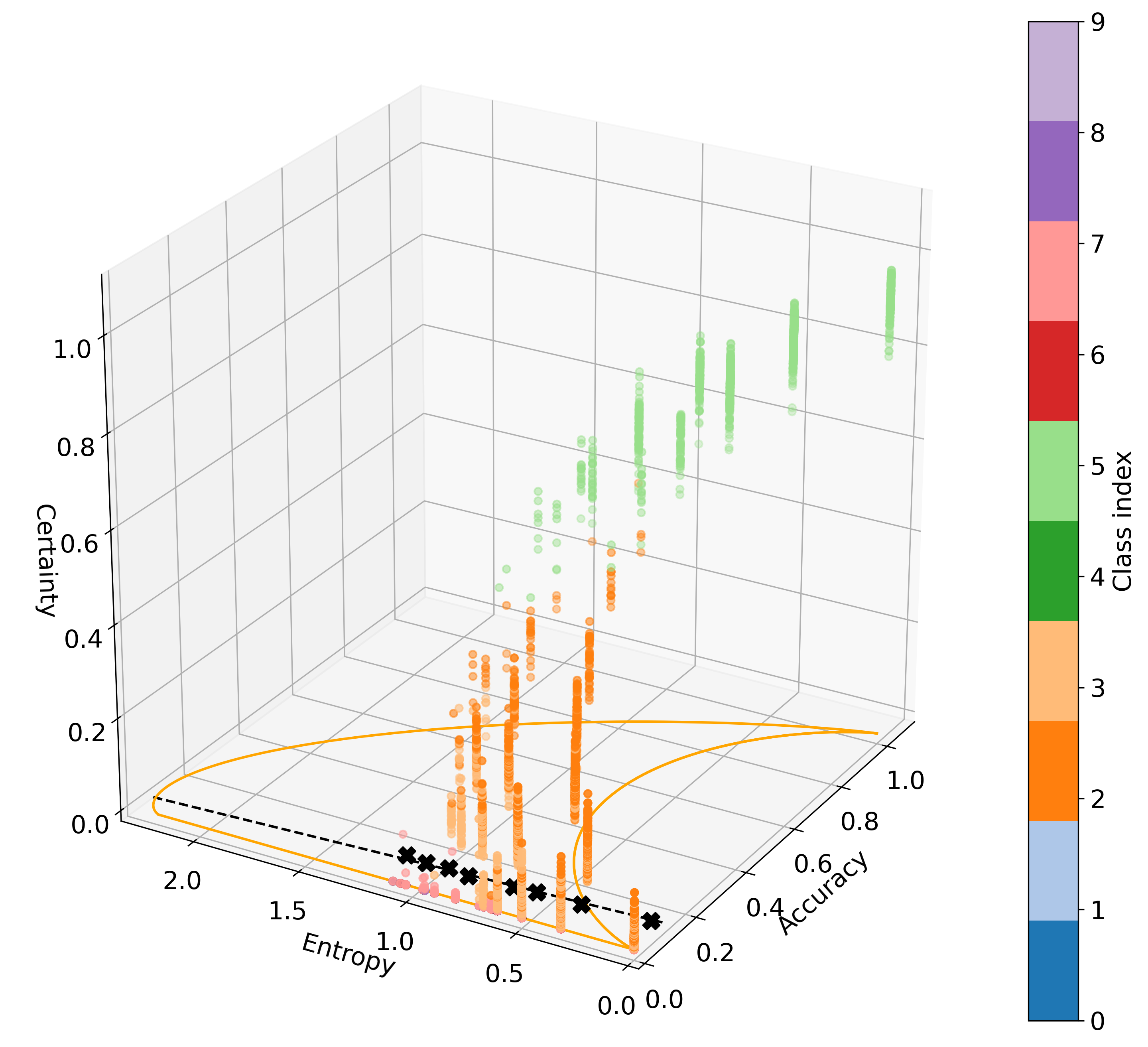}
        \caption{Image perturbation with $\sigma = 10.0$}
        \label{fig:image_logistic_10}
    \end{subfigure}
    \caption{Weight and Image Perturbation Evaluated on Na\"ive multinomial Regression Model}
    \label{fig:weight_logistic}
\end{figure}

Looking at Figure~\ref{fig:weight_logistic_0.1} and Figure~\ref{fig:weight_logistic_10}, it is evident that a higher value for sigma increases the overall entropy of the model. The model also tends to move towards the lowest expected accuracy, $10\%$ in this case, since a high entropy model can be seen as truly random. We also see that the regression does have a negative slope, although weaker for higher sigma. It is also clear that the \textit{mass} of the predictions have moved towards higher entropy and lower accuracy.

Looking at Figure~\ref{fig:image_logistic_0.1} and Figure~\ref{fig:image_logistic_10}, another pattern is seen. At high image perturbation levels, only a few classes are visible, while most are clustered at $0\%$ accuracy, while for lower perturbation it is more diverse and similar to weight perturbation. The negative slope for the regression line is strongly present for low sigma. But for high sigma we have lost this correlation entirely. The mass has also significantly moved towards a lower accuracy, although while the entropy has stayed the same.

\subsection{Pretrained ConvNeXt}
\begin{table}[H]
    \caption{PSI and PI Metrics for Pretrained ConvNeXt}
    \centering
    \begin{tabular}{| c | c c c c | c c c c | c | c |}
        \hline
        \multirow{2}{*}{\diagbox{$\sigma$}{$\lambda$}} & \multicolumn{4}{c|}{$\psi_w$} & \multicolumn{4}{c|}{$\psi_i$} & \multirow{2}{*}{$\pi_w$} & \multirow{2}{*}{$\pi_i$}\\
        & 0.1 & 0.5 & 1 & 2 & 0.1 & 0.5 & 1 & 2 & & \\
        \hline
        0.1 & 0.100 & 0.100 & 0.101 & 0.102 & 0.705 & 0.870 & 1.077 & 1.492 & 0.838 & 0.276 \\
        0.5 & 0.100 & 0.102 & 0.104 & 0.109 & 0.303 & 0.437 & 0.605 & 0.940 & 0.839 & 0.668 \\
          1 & 0.100 & 0.100 & 0.100 & 0.099 & 0.103 & 0.106 & 0.109 & 0.115 & 0.835 & 0.836 \\
         10 & 0.099 & 0.103 & 0.106 & 0.114 & 0.100 & 0.104 & 0.110 & 0.123 & 0.839 & 0.841 \\
        \hline
    \end{tabular}
    \label{tab:pretrained_convnext}
\end{table}

Looking at Table~\ref{tab:pretrained_convnext} we can observe that in the weight perturbation case, any perturbation level, low or high, affects the metrics heavily. However, in the image perturbation case, it is a more linear correlation.

\begin{figure}[H]
    \centering
    \begin{subfigure}[b]{0.4\textwidth}
        \centering
        \includegraphics[width=\textwidth]{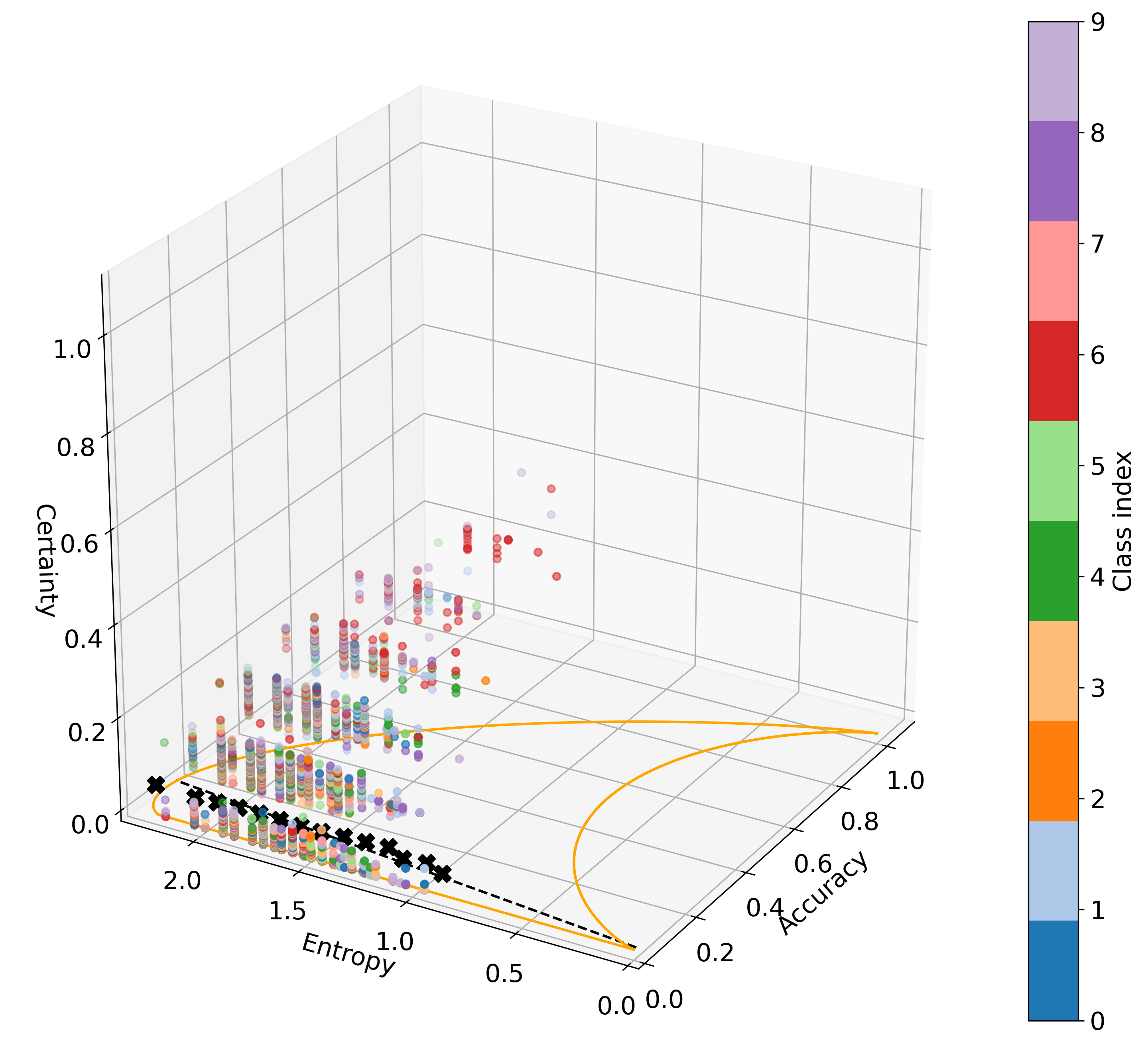}
        \caption{Weight perturbation with $\sigma = 0.1$}
        \label{fig:weight_pretrained_convnext_0.1}
    \end{subfigure}
    \hfill
    \begin{subfigure}[b]{0.4\textwidth}
        \centering
        \includegraphics[width=\textwidth]{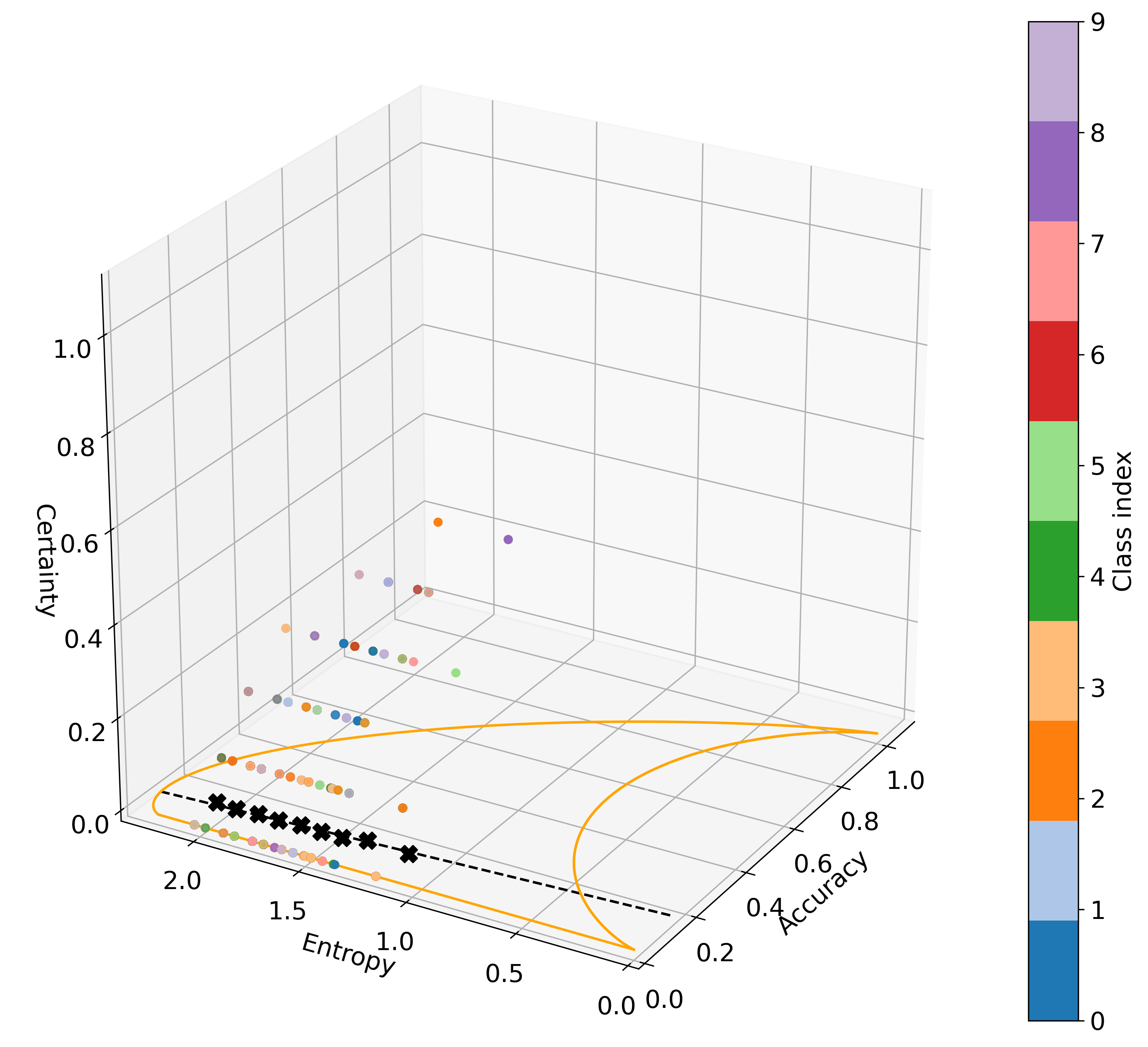}
        \caption{Weight perturbation with $\sigma = 10.0$}
        \label{fig:weight_pretrained_convnext_10}
    \end{subfigure}

    \begin{subfigure}[b]{0.4\textwidth}
        \centering
        \includegraphics[width=\textwidth]{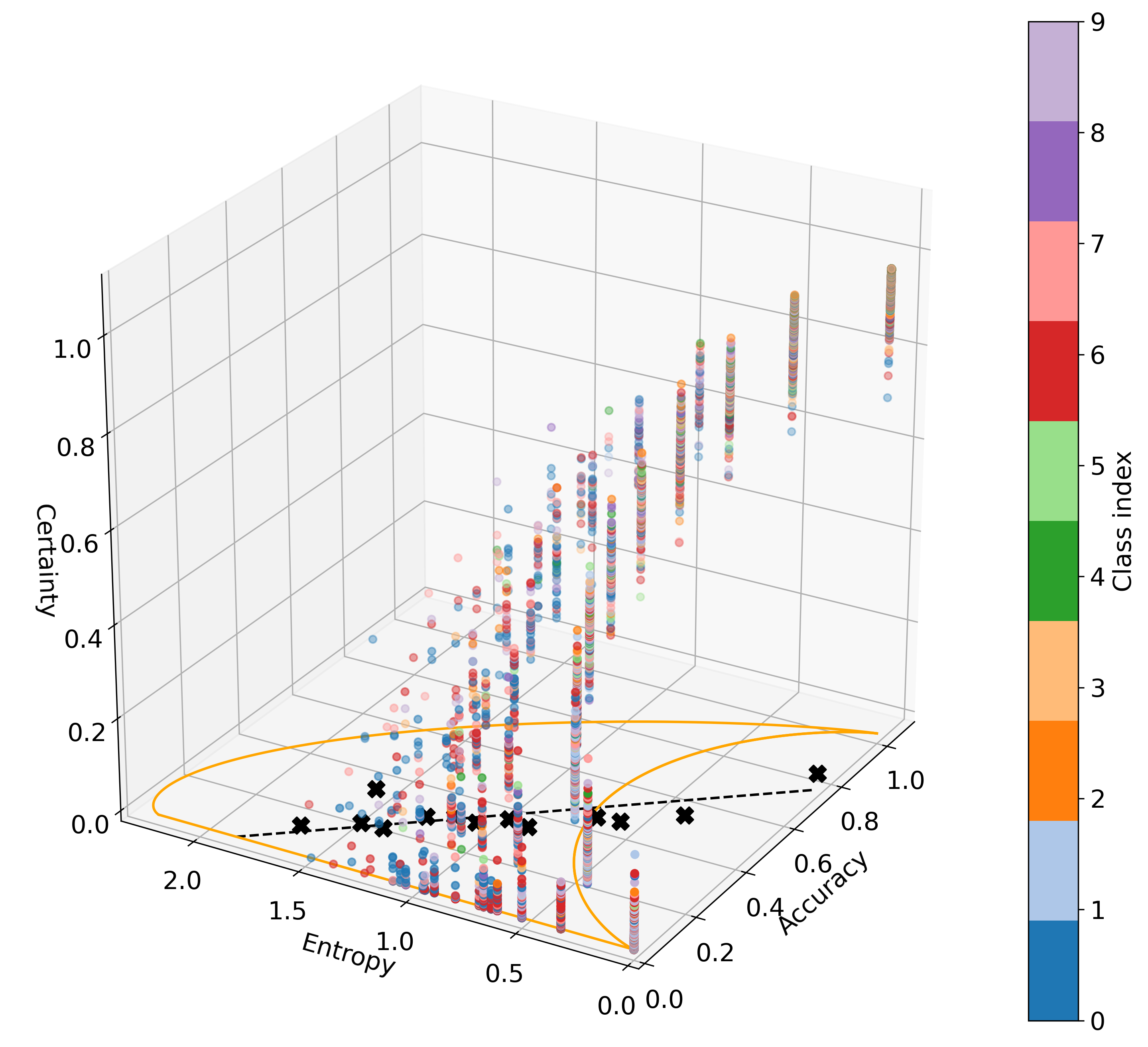}
        \caption{Image perturbation with $\sigma = 0.1$}
        \label{fig:image_pretrained_convnext_0.1}
    \end{subfigure}
    \hfill
    \begin{subfigure}[b]{0.4\textwidth}
        \centering
        \includegraphics[width=\textwidth]{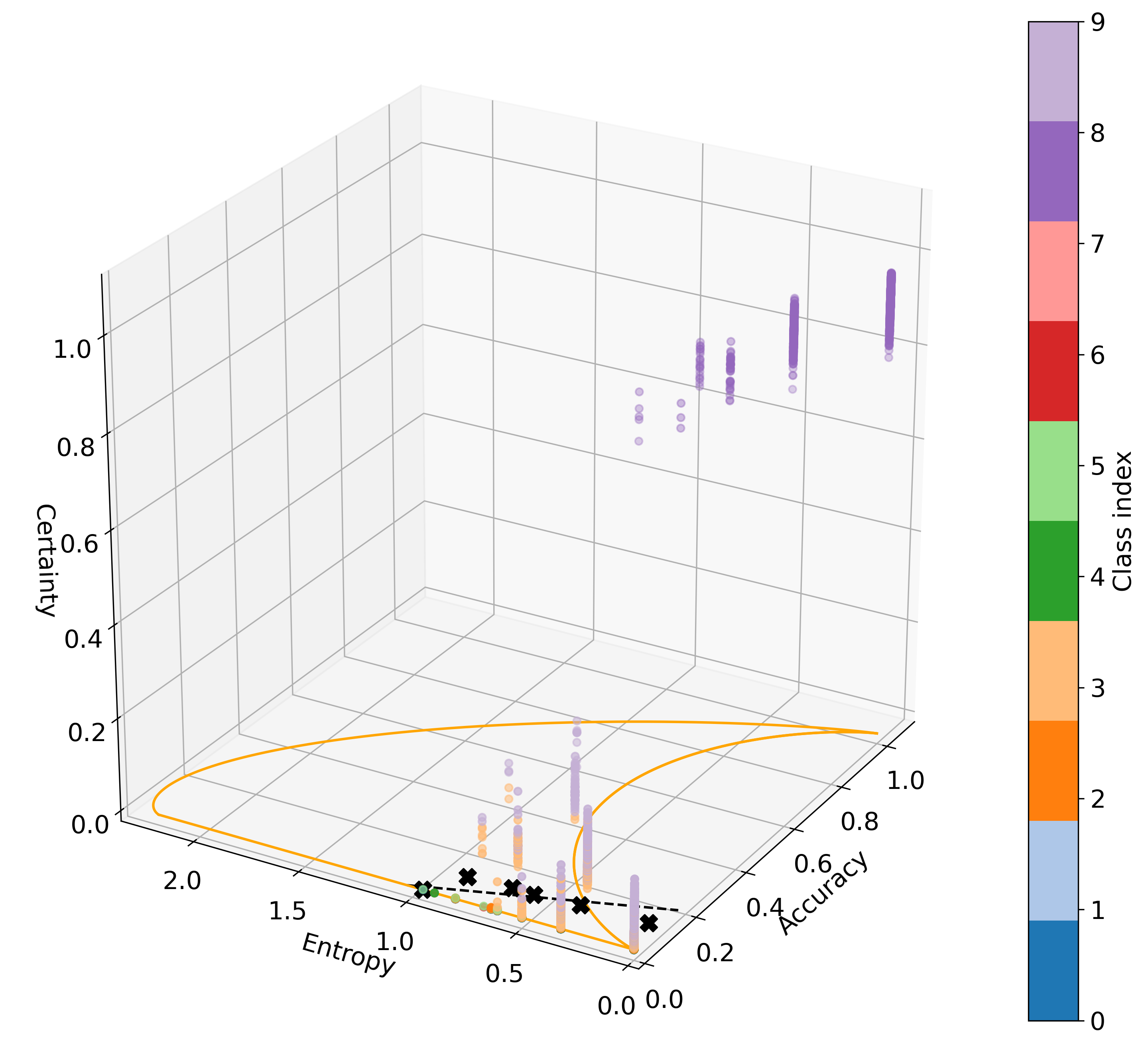}
        \caption{Image perturbation with $\sigma = 10.0$}
        \label{fig:image_pretrained_convnext_10}
    \end{subfigure}
    \caption{Weight and Image Perturbation Evaluated on Pretrained ConvNeXt Model}
    \label{fig:weight_pretrained_convnext}
\end{figure}

From looking at Figure~\ref{fig:weight_pretrained_convnext_0.1} and Figure~\ref{fig:weight_pretrained_convnext_10} we see the model's predictions tend towards maximum entropy. The biggest difference between the two perturbation levels is in the spread of certainty, with higher perturbation, the certainty drastically decreases. In neither case does the mass of the predictions or the regression line change in any way.

Looking at Figure~\ref{fig:image_pretrained_convnext_0.1} and Figure~\ref{fig:image_pretrained_convnext_10}, we observe the same pattern seen previously in the image perturbation case. We again observe that the regression line for a low sigma is present and has a negative slope. Interestingly, the mass has moved to a lower overall entropy, while sacrificing accuracy.

\subsection{Pretrained ViT}
\begin{table}[H]
    \caption{PSI and PI Metrics for Pretrained ViT}
    \centering
    \begin{tabular}{| c | c c c c | c c c c | c |c|}
        \hline
        \multirow{2}{*}{\diagbox{$\sigma$}{$\lambda$}} & \multicolumn{4}{c|}{$\psi_w$} & \multicolumn{4}{c|}{$\psi_i$} & \multirow{2}{*}{$\pi_w$} & \multirow{2}{*}{$\pi_i$}\\
        & 0.1 & 0.5 & 1 & 2 & 0.1 & 0.5 & 1 & 2 & &\\
        \hline
        0.1 & 0.097 & 0.091 & 0.085 & 0.071 & 0.625 & 0.754 & 0.915 & 1.236 & 0.858 & 0.354 \\
        0.5 & 0.097 & 0.085 & 0.070 & 0.049 & 0.522 & 0.711 & 0.948 & 1.421 & 0.848 & 0.471 \\
          1 & 0.099 & 0.098 & 0.096 & 0.094 & 0.177 & 0.231 & 0.299 & 0.435 & 0.848 & 0.784 \\ 
         10 & 0.100 & 0.101 & 0.104 & 0.108 & 0.096 & 0.096 & 0.096 & 0.096 & 0.850 & 0.852 \\
        \hline
    \end{tabular}
    \label{tab:pretrained_vit}
\end{table}

From Table~\ref{tab:pretrained_vit} we observe the same pattern as with the ConvNeXt model. The model is heavily affected by weight perturbation, at any level, whilst being less affected by image perturbation.

\begin{figure}[H]
    \centering
    \begin{subfigure}[b]{0.4\textwidth}
        \centering
        \includegraphics[width=\textwidth]{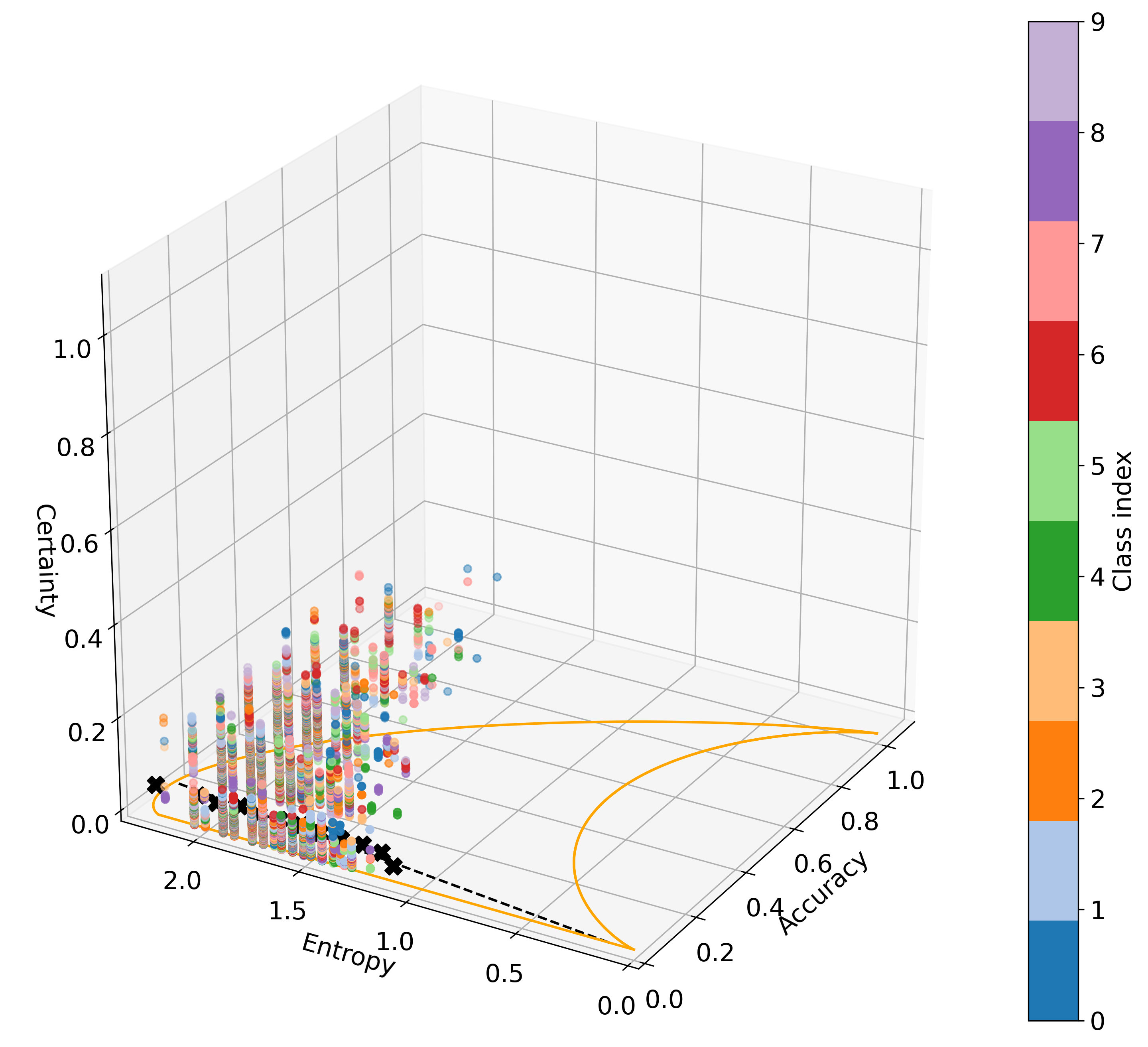}
        \caption{Weight perturbation with $\sigma = 0.1$}
        \label{fig:weight_pretrained_vit_0.1}
    \end{subfigure}
    \hfill
    \begin{subfigure}[b]{0.4\textwidth}
        \centering
        \includegraphics[width=\textwidth]{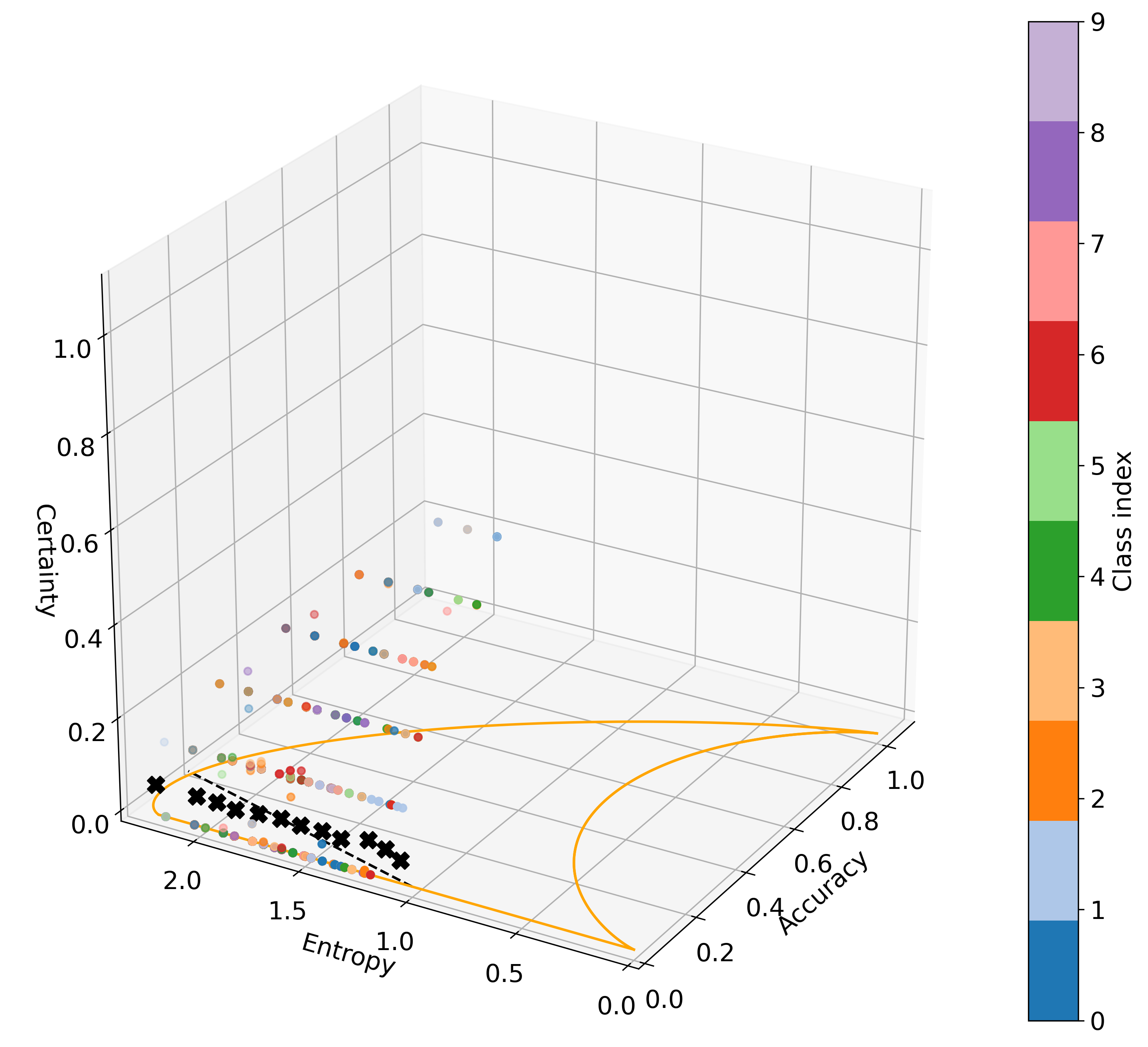}
        \caption{Weight perturbation with $\sigma = 10.0$}
        \label{fig:weight_pretrained_vit_10}
    \end{subfigure}
    
    \begin{subfigure}[b]{0.4\textwidth}
        \centering
        \includegraphics[width=\textwidth]{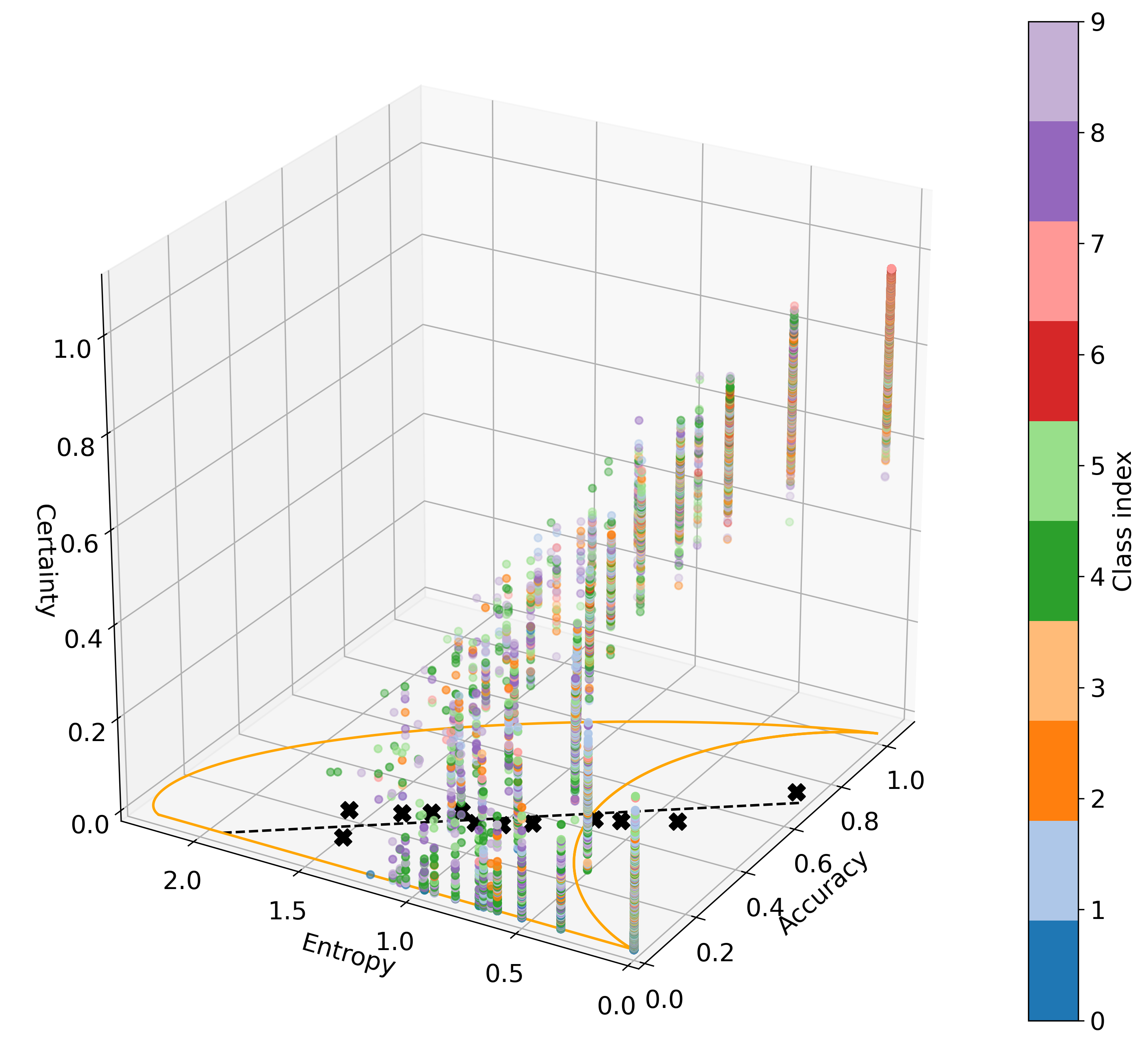}
        \caption{Image perturbation with $\sigma = 0.1$}
        \label{fig:image_pretrained_vit_0.1}
    \end{subfigure}
    \hfill
    \begin{subfigure}[b]{0.4\textwidth}
        \centering
        \includegraphics[width=\textwidth]{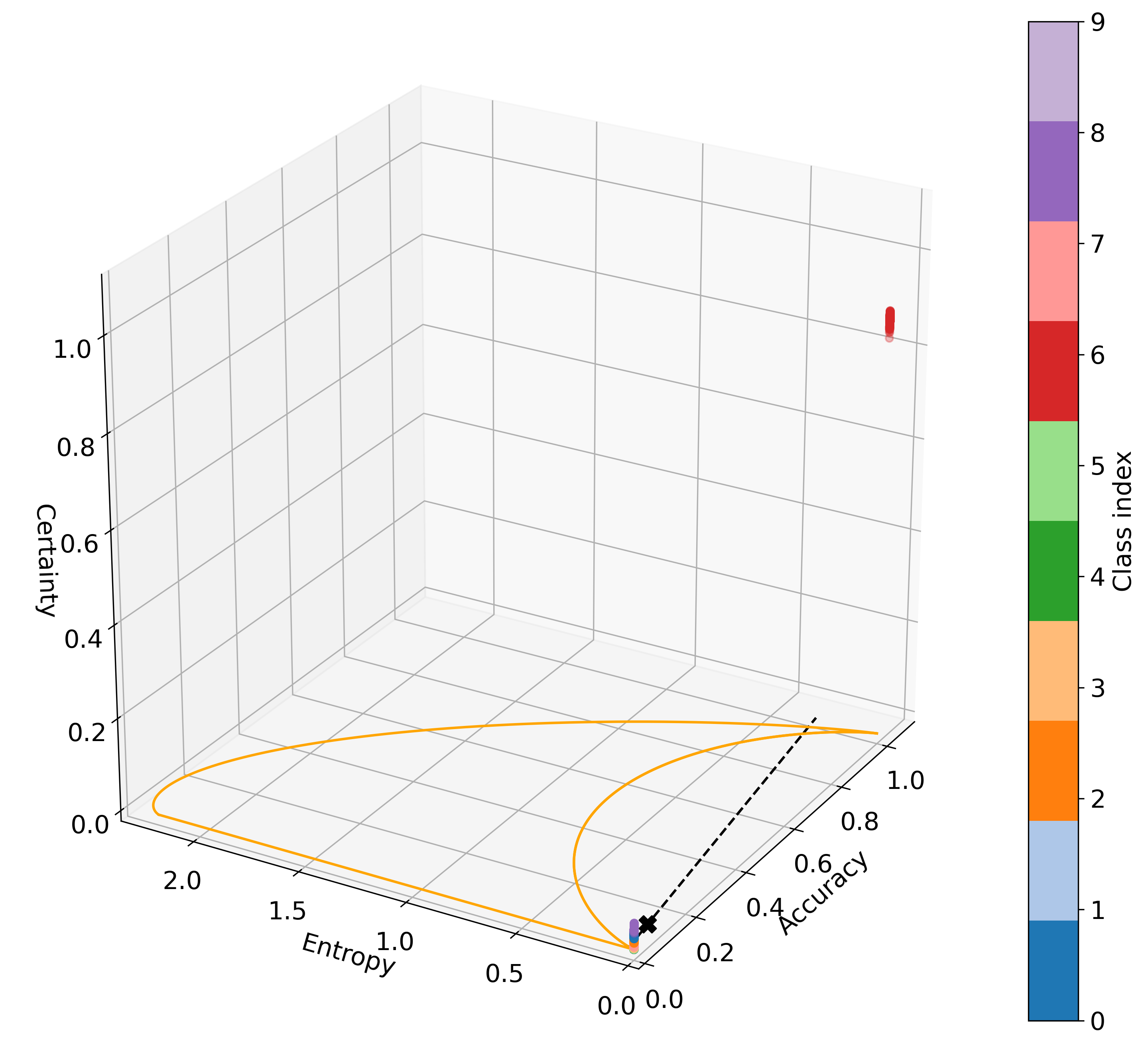}
        \caption{image perturbation with $\sigma = 10.0$}
        \label{fig:image_pretrained_vit_10}
    \end{subfigure}
    \caption{Weight and Image Perturbation Evaluated on Pretrained ViT}
    \label{fig:weight_pretrained_vit}
\end{figure}

Looking at Figure~\ref{fig:weight_pretrained_vit_0.1} and Figure~\ref{fig:weight_pretrained_vit_10} we see that this model has a much higher overall entropy, and the spread of the certainty at lower perturbation levels being large. The case for higher weight perturbation is still continuous. As seen in the ConvNeXt model, the mass and regression line stay intact in both cases for weight perturbation.

Observing Figure~\ref{fig:image_pretrained_vit_0.1} and Figure~\ref{fig:image_pretrained_vit_10} a new extreme pattern is seen. In previous high image perturbation, only a selection of classes were visible, the ViT only has one class visible. As previously, the image perturbation for low sigma shows the expected negative slope. However, interestingly, for sigma $10.0$ the entire mass is close to $0$ entropy and therefore the accuracy only being $100\%$ or $0\%$.

%% file: sections/Discussion/discussion.tex
As seen in the results, when high perturbation is injected into the image, the mass of the predictions ends up being both low in entropy and accuracy. However, the only predictions that escape this pit is only a handful, both in mass and class index. A possible explanation is that, due to extreme perturbation levels the image can be seen as arbitrary, the models' prediction is a \textit{guess}. We can also view these \textit{guesses} as fallback predictions, when the model is given an arbitrary input it falls back to these classes since no inherent meaning is found.

Looking at the weight perturbation graphs, especially when the perturbation level is high, there is a great spread of certainty in the predictions.
One likely cause could be because the output from the last layer of the networks before the softmax activation has values blown out of proportion, which leads to the softmax output containing all zeroes except a singular $1$. This leads to the certainty being on a specific level which is the number of predictions for that class divided by the number of times that data point has been sampled. In our case, it would be $\hat{y}_{\sigma, Y}/n$ where $n$ is the number of iterations, which was 10, hence $\hat{y}_{\sigma, Y}/10$. However, this does not seem to be the case on Na\"ive multinomial Regression. Our reasoning for this behavior is that the Na\"ive Multinomial model is considerably smaller than the other models.

Comparing the results for image perturbation compared to weight perturbation, we clearly see that all models perform better on image perturbation. We reason that this is due to perturbing the image, creates magnitudes smaller overall perturbation than perturbing every weight in the entire network. As discussed previously, if we perturb the entire network, the output could be blown out of proportion, whilst the image perturbation still yields a numerical valid prediction. Also, as previously discussed, the overall complexity of the model matters. If the model is complex and deep, there will be more trainable parameters, therefore, the perturbed model's output per layer will propagate and accumulate to the final output. 

Finally, referring back to our proposed theoretical framework and method (Section~\ref{sec:entropy_framework}) and using our found results, we can see a clear correlation between our proposed theory and our practical results. 

\subsection{Future Improvements} \label{sec:future_improvements}
One easy way to improve the overall results of this study is simply to improve the statistical inference by increasing the number of samples ($n$). For further improvements on this study, we would like to see how perturbation on specific areas of the models (for example, only the classification head or the feature extraction part of a network) would affect the outputs. Along with using a bigger sample of models of varying types and trained to different levels (for example, having trained a ''bad'', ''decent'' and ''good'' version of the same model) and see if the assumptions still hold. Also examining the effects on different model sizes, such as the ConvNeXt Tiny, Base, or XL, assessing if the PSI metric changes with more parameters.

\subsection{Conclusions}\label{sec:conclusions}
    To conclude this study, the questions presented in Section  \ref{sec:objectives} will be answered:
    \begin{enumerate}[wide, labelwidth=!, labelindent=0pt]
        \item Can entropy be used to find methods for quantifying the uncertainty of a machine learning model?\\
        Yes, the results of this study show that Shannon entropy does provide a theoretical framework to quantify uncertainty. 
         
        \item If this is possible, in what ways can these methods be applied?\\
        The study explored the  entropy-based metrics PI and PSI across the different models: Naïve Multinomial Regression, ConvNeXt, and ViT under varying levels of perturbations. The EAC (entropy-accuracy-certainty) calculations and graphs also gave us important insights.
        
        \item Does this work equally well on all models?\\
        No, as discussed in the earlier section the effectiveness and evaluation differed a lot across the models. For more complex and sophisticated models, even lower sigmas are needed to be analyzed. 
    \end{enumerate}

%% file: sections/Ethics/Ethics.tex
As previously noted, this research focuses on enhancing the reliability of computer vision models to make accurate predictions. But what practical applications might this have? 

Image recognition technology is increasingly common, utilized across various industries and by individuals daily. Some examples that utilize this technology include security systems, autonomous vehicles, healthcare diagnostics, and aerial surveillance. By utilizing our metrics and methods, one could evaluate how the model would behave on unseen data, before knowing the ground truth. To further evaluate the confidence of the model's prediction in these areas, less uncertain predictions can be made, which improves the overall prediction.



\subsection{Accountability and Impact on Decision Making}
A frequent question arises: Who is responsible for the decisions made by models? As mentioned in the introduction, the design of a neural network is inspired by the brain's functionality—it needs to learn, adapt, and make decisions. Like humans, who can make minor or major mistakes and be held accountable for their actions, the question becomes, how can a machine be held accountable?

In addressing this issue, it seems reasonable to hold the engineer who developed the technology responsible. However, does this imply that the developer made errors intentionally or out of ignorance?

Organizations like the World Federation of Engineering Organizations \cite{WFEO2023} and the Swedish organization "Sveriges Ingenjörer" \cite{SverigesIngenjorer2024} have established a code of ethics that engineers must adhere to. This code suggests that engineers are responsible for their work, should avoid employment with companies of dubious ethics, and strive to perform to the best of their abilities.

Given these ethical guidelines, one might assume that an engineer would do their utmost to develop techniques that adhere to these standards. However, engineers often are not the only individuals involved; companies and operators, among others, also play significant roles. This complicates the question of responsibility: what if external factors or disturbances affected the technique? While it remains unclear exactly who is to blame in such scenarios, it is clear that if all parties involved adhere to ethical standards and fulfill their responsibilities, the risk of this occuring is significantly decrease.

When discussing computers that can make decisions on our behalf, it's important to consider the influence these decisions have on our choices and how we might use or apply them. For instance, models in security cameras, how much should we trust a machine's decisions? What happens if the machine makes an error without the operators noticing?

This research requires further development before it can be effectively implemented in real-world scenarios. No solution is flawless, however, it is essential to acknowledge the potential for mistakes and errors. A neuroscience study from UCL \cite{NeuroscienceNews2024} highlighted that perceptual decisions could be skewed by the effort involved in an action, suggesting that humans often choose the path of least resistance when making decisions. These models represent that path. Therefore, it is crucial for users to assess the information provided and consider the possible uncertainties critically.





%% file: appendix/C.tex
The source code for this study is available at \href{https://www.github.com/rezaarezvan/MVEX11}{GitHub} under the MIT license.